\documentclass{article}

\usepackage[main, final]{neurips_2026}

\usepackage[utf8]{inputenc} 
\usepackage[T1]{fontenc}    
\usepackage{hyperref}       
\usepackage{url}            
\usepackage{booktabs}       
\usepackage{amsfonts}       
\usepackage{nicefrac}       
\usepackage{microtype}      
\usepackage{xcolor}         
\usepackage{amsmath}        
\usepackage{multirow}       
\usepackage{graphicx}       
\usepackage{amsthm}
\usepackage{enumitem}
\usepackage{placeins}

\usepackage[utf8]{inputenc} 
\usepackage[T1]{fontenc}    
\usepackage{hyperref}       
\usepackage{url}            
\usepackage{booktabs}       
\usepackage{array}          
\usepackage{graphicx}       
\usepackage{pifont}
\usepackage{amsfonts}       
\usepackage{microtype}      
\usepackage{xcolor}         

\usepackage{booktabs}
\usepackage{color}
\usepackage{xcolor}
\usepackage{amssymb}
\usepackage{amsmath}

\usepackage{amsmath}        

\newcolumntype{L}[1]{>{\raggedright\arraybackslash}p{#1}}
\newcolumntype{C}[1]{>{\centering\arraybackslash}p{#1}}
\newcolumntype{R}[1]{>{\raggedleft\arraybackslash}p{#1}}

\definecolor{purple}{RGB}{128, 0, 128}

\makeatletter
\g@addto@macro\normalsize{%
  \setlength{\abovedisplayskip}{4pt plus 1pt minus 1pt}%
  \setlength{\belowdisplayskip}{4pt plus 1pt minus 1pt}%
  \setlength{\abovedisplayshortskip}{2pt plus 1pt minus 1pt}%
  \setlength{\belowdisplayshortskip}{3pt plus 1pt minus 1pt}%
  \setlength{\jot}{2pt}%
}
\makeatother
\normalsize

\title{Differencing the Diffusion Trajectory toward Uncertain Components for Time Series Forecasting}

\author{%
  {Chen Su\textsuperscript{\ding{171}}, \quad Yuanhe Tian\textsuperscript{\ding{170}}, \quad Yan Song\textsuperscript{\ding{171}}\thanks{Corresponding author.}}\\
  {\textsuperscript{\ding{171}}University of Science and Technology of China \quad \textsuperscript{\ding{170}}Zhongguancun Academy}\\
  {\textsuperscript{\ding{171}}\texttt{suchen4565@mail.ustc.edu.cn} \quad
  \textsuperscript{\ding{170}}\texttt{yhtian94@gmail.com} \quad
  \textsuperscript{\ding{171}}\texttt{clksong@gmail.com}}
}

\begin{document}

\maketitle

\begin{abstract}
Diffusion models have become a widely used framework for probabilistic time series forecasting, modeling the distribution of future values given an observed history. In time series forecasting, however, the future continues the observed history, creating an asymmetry the standard diffusion process leaves unaddressed, with slowly-varying content largely determined by the observed continuity while higher-frequency dynamics carry most of the residual uncertainty. Existing diffusion-based forecasters decouple this asymmetry through an external rule before generation, leaving the corruption trajectory blind to which parts of the target the history can already anchor. We propose \textsc{DiffDiff}, a diffusion framework that embeds this predictability asymmetry into the diffusion trajectory itself, so that a single end-to-end diffusion process becomes aware of which parts of the target the history can already anchor. \textsc{DiffDiff} makes the forward operator step-dependent so that the noisy intermediate state progressively shifts from the target itself toward its second-order differenced structure, while a conditioning pathway supplies the denoiser with both value-domain and differential history information balanced by a stage-adaptive gate at each diffusion step. The terminal distribution approaches a standard Gaussian, preserving compatibility with existing samplers. {On seven benchmarks across four prediction horizons, \textsc{DiffDiff} outperforms six diffusion baselines, while our analyses show that its trajectory reallocates modeling effort according to history predictability and improves distributional fidelity.}\footnote{Code is released at {\url{https://github.com/synlp/DiffDiff}}.}
\end{abstract}

\section{Introduction}
\label{sec:intro}

Probabilistic time series forecasting aims to predict the conditional distribution of future values from a historical observation window, providing both accurate point estimates and calibrated uncertainty quantification.
Reliable distributional forecasts are essential in risk-sensitive applications such as energy dispatch~\cite{nowotarski2018electricity,chou2018energy}, financial hedging~\cite{cao2019financial,dingli2017financial}, and extreme weather preparedness~\cite{hewage2020weather,karevan2020weather}, where decisions depend not only on expected outcomes but also on the range of plausible futures.
Denoising diffusion probabilistic models (DDPMs)~\cite{sohl2015deep,ho2020denoising,song2021sde,rombach2022ldm,peebles2023dit,gan2023idesigner,wu2024taiyi,lu2022dpmsolver} have recently emerged as a powerful generative framework for this task because they model complex predictive distributions without restrictive parametric assumptions \cite{su2025diffusion}.

Within the forecasting target, however, predictive uncertainty is not uniformly distributed.
Components that remain strongly correlated with the observed history, such as smooth trends, tend to carry little remaining uncertainty, whereas the components the history under-determines, such as rapid fluctuations beyond what can be extrapolated from the past, concentrate the residual uncertainty and are where a generative model's capacity is best spent~\cite{li2025d3u,wang2025falda}.
Recent diffusion-based forecasters have begun to exploit this asymmetry by decoupling the target into a deterministic component and an uncertain component before any diffusion modeling, applying diffusion only to the latter~\cite{li2025d3u,zhang2024cdpm}.
Such external decoupling validates that this heterogeneity matters for forecasting, but it also localizes the asymmetry outside the diffusion pipeline, leaving the diffusion process itself task-agnostic to the structure it is meant to model.
Modifications to the diffusion process itself have also been explored within forecasting pipelines, such as introducing learnable, data-aware components into the forward process~\cite{li2024tmdm,rishi2025cndiff}, matching the diffusion endpoint variance to the data's non-stationary variation~\cite{ye2025nsdiff}, or slowing the decay of low-frequency content along the diffusion trajectory~\cite{wang2025matsd}.
Yet none of these designs reshapes what the trajectory preserves of the target so that intermediate states emphasize the parts the history cannot already supply.
The diffusion's capacity is therefore spent recovering both the parts the history could already anchor and those it cannot, instead of being focused where forecasting uncertainty actually concentrates.

We propose \textsc{DiffDiff} (\textit{Differencing Diffusion}), a forecasting diffusion framework that embeds this asymmetry directly into the diffusion trajectory so that the diffusion process itself becomes aware of which parts of the target the history can already anchor, rather than offloading the asymmetry to an external decomposition.
The forward operator is made step-dependent so that, as diffusion progresses, components readily reproducible from history are progressively suppressed in the noisy state, while components the history under-determines remain salient and absorb most of the trajectory's representational budget.
To compensate for this suppression, the conditioning pathway combines a value-domain encoding of the observed history with a differencing branch that captures its temporal evolution, and a stage-adaptive gate then re-supplies the suppressed content to the denoiser at the diffusion stages where the noisy target alone is no longer informative.
The diffusion process thus remains a single end-to-end estimator over the full target, while its modeling capacity is reallocated along the trajectory, with history-anchored content sustained through conditioning and diffusion concentrating on components for which the history is least informative.
Although the intermediate states are reshaped, the terminal distribution approaches a standard Gaussian, preserving compatibility with existing noise schedules and accelerated samplers.

\vspace{-0.2cm}
\section{Related Work}
\label{sec:related-work}
\vspace{-0.1cm}
\subsection{Diffusion Models for Time Series Forecasting}
\label{sec:rw-tsd}

Diffusion-based forecasting evolved from autoregressive designs~\cite{rasul2021timegrad,liu2024stochdiff} to non-autoregressive successors~\cite{tashiro2021csdi,alcaraz2023sssd,shen2023timediff,kollovieh2024tsdiff,su2025multimodal} that avoid error accumulation.
Subsequent work enriched the conditioning pathway through multi-resolution~\cite{shen2024mrdiff,fan2024mgtsd}, seasonal-trend~\cite{yuan2024diffusionts}, variational autoencoder (VAE)-hybrid~\cite{li2022d3vae}, and retrieval-augmented~\cite{liu2024ratd,blattmann2022rdm,jing2022retrievaltsf,su2025fusing} designs.
For multimodal forecasting, there are studies enhance textual inputs through reinforcement learning, with rewards based on forecasting performance and task relevance \cite{su2025text}.
Components strongly correlated with the history are recoverable by deterministic predictors~\cite{nie2023patchtst,liu2024itransformer,wu2023timesnet,oreshkin2020nbeats,liu2022nonstationary,zhou2022fedformer,zhang2023crossformer,liu2021pyraformer}, while the residual carries most of the predictive entropy. D3U~\cite{li2025d3u} freezes a point predictor over trend and seasonality and CDPM~\cite{zhang2024cdpm} routes the trend through a polynomial module, in each case pre-cleaning the residual by a fixed external rule that locks the deterministic-uncertain boundary before diffusion runs.
A complementary strand modifies the diffusion process itself end-to-end. TMDM~\cite{li2024tmdm} and CN-Diff~\cite{rishi2025cndiff} reshape the terminal prior, with TMDM centering it on a transformer point forecast and CN-Diff training a learned multilayer perceptron (MLP) forward transformation jointly with a conditional prior.
NsDiff~\cite{ye2025nsdiff} learns a history-conditional per-cell diagonal noise variance, but the reshaping acts only on the noise while leaving the target encoding structurally fixed.
MA-TSD~\cite{wang2025matsd} adopts a non-isotropic forward operator built from progressive moving-average smoothing that preserves low-frequency content across the chain.
\textsc{DiffDiff} likewise reshapes what the intermediate state encodes, but redirects the trajectory's budget toward components the history under-determines while the conditioning pathway adaptively re-supplies the suppressed content. The asymmetry between deterministic and uncertain content thus emerges along the trajectory itself rather than being preset by an external decomposition.

\subsection{Non-Isotropic Diffusion Processes}
\label{sec:rw-niso}

Several studies have explored diffusion models with non-isotropic forward processes in domains beyond time series \cite{tian2024diffusion,su2026learning}.
Blurring Diffusion~\cite{hoogeboom2023blurring} and Inverse Heat Dissipation~\cite{rissanen2023heat} replace isotropic Gaussian corruption with discrete cosine transform (DCT)-domain frequency decay and heat-equation partial differential equation (PDE) dynamics, respectively, for image generation \cite{jin2024improving}.
Soft Diffusion~\cite{daras2023soft} formulates a general training objective for arbitrary linear corruptions combined with Gaussian noise, while Cold Diffusion~\cite{bansal2023cold} removes the Gaussian noise entirely and inverts arbitrary deterministic degradations such as blurring and masking.
Whitened Score Diffusion~\cite{alido2025whitened} enables arbitrary Gaussian forward processes with non-identity covariance via whitened score matching, targeting imaging inverse problems.
\textsc{DiffDiff} specializes this non-isotropic framework for forecasting, where the operator is shaped by the within-target predictability heterogeneity inherent to forecasting targets rather than by image-domain priors such as blur or frequency decay.

\vspace{-0.2cm}
\section{Method}
\label{sec:method}
\begin{figure}[t]
\centering
\includegraphics[width=\textwidth, trim=0 10 0 0]{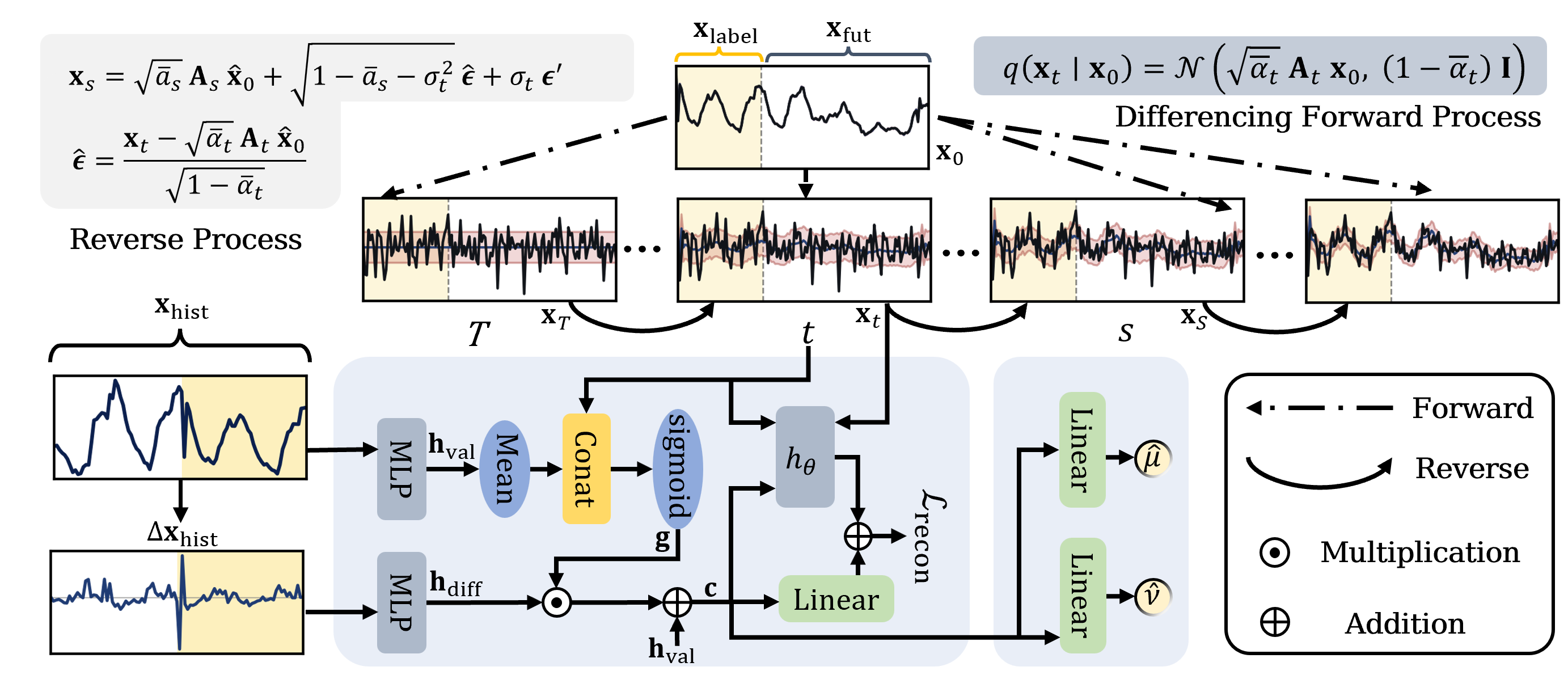}
\caption{Overall architecture of \textsc{DiffDiff}. \textbf{Top}: the differencing forward process gradually shifts the clean target $\mathbf{x}_0$ toward its differenced structure under the step-dependent operator $\mathbf{A}_t$, and the deterministic reverse step recovers $\mathbf{x}_s$ from the denoiser prediction $\hat{\mathbf{x}}_0$. \textbf{Bottom}: the stage-adaptive conditional denoiser encodes both the value-domain history $\mathbf{x}_{\text{hist}}$ and its first-order differences $\Delta\mathbf{x}_{\text{hist}}$, fuses them through a sigmoid gate conditioned on the diffusion timestep, and outputs the value-domain prediction along with auxiliary statistics for instance denormalization.}
\vspace{-0.2cm}
\label{fig:model}
\end{figure}

\vspace{-0.1cm}
Given an observed history window $\mathbf{x}_{\text{hist}}$, probabilistic time series forecasting aims to model the conditional distribution $p(\mathbf{x}_{\text{fut}} \mid \mathbf{x}_{\text{hist}})$ of a future window $\mathbf{x}_{\text{fut}}$. To learn this distribution with diffusion modeling, we denote the clean forecasting target by $\mathbf{x}_0$ and define a forward process $q(\mathbf{x}_t \mid \mathbf{x}_0)$ that progressively transforms $\mathbf{x}_0$ into noisy latent states $\mathbf{x}_t$ over diffusion steps $t = 1, \ldots, T$, with the terminal state $\mathbf{x}_T$ approaching a standard Gaussian prior $\mathcal{N}(\mathbf{0}, \mathbf{I})$. A parameterized reverse process $p_\theta(\mathbf{x}_{t-1} \mid \mathbf{x}_t, \mathbf{x}_{\text{hist}})$, implemented by a denoising network $f_\theta(\mathbf{x}_t, t, \mathbf{x}_{\text{hist}})$, learns to invert this transformation, so that iteratively applying the reverse transitions from a terminal sample yields a draw from $p_\theta(\mathbf{x}_0 \mid \mathbf{x}_{\text{hist}})$. As depicted in Figure~\ref{fig:model}, \textsc{DiffDiff} couples a step-dependent forward process that gradually encodes differential structure into the noisy state with a stage-adaptive conditional denoiser that fuses value-domain and differential history information.

\vspace{-0.2cm}
\subsection{Forward Process with Progressive Differencing}
\label{sec:forward}

The information that a denoiser must recover from a noisy intermediate state depends on what signal content that state preserves. We design the forward process so that, as diffusion progresses, the noisy intermediate state retains less of the components readily reproducible from the observed history and more of the components the history under-determines. The diffusion's modeling capacity is thereby aligned with the predictability heterogeneity within the forecasting target rather than spread uniformly across it. {Appendix~\ref{sec:appendix-forecasting-difficulty} further shows that rapid local changes tend to coincide with greater forecasting difficulty.}
Given a clean target sequence $\mathbf{x}_0$, we define a generalized forward marginal
\begin{equation}
q(\mathbf{x}_t \mid \mathbf{x}_0) = \mathcal{N}\!\left(\sqrt{\bar{\alpha}_t}\,\mathbf{A}_t\,\mathbf{x}_0,\;(1 - \bar{\alpha}_t)\,\mathbf{I}\right)
\setlength{\abovedisplayskip}{5pt}
\setlength{\belowdisplayskip}{5pt}
\label{eq:forward-marginal}
\end{equation}
where $\bar{\alpha}_t = \prod_{\tau=1}^{t}(1-\beta_\tau)$ is the cumulative noise coefficient following a cosine schedule~\cite{nichol2021improved} with $\bar{\alpha}_0 = 1$ and $\bar{\alpha}_T \approx 0$, and $\mathbf{A}_t$ is a step-dependent transition matrix that determines what signal representation is preserved at diffusion step $t$. Equivalently, by the reparameterization trick,
\begin{equation}
\mathbf{x}_t = \sqrt{\bar{\alpha}_t}\,\mathbf{A}_t\,\mathbf{x}_0 + \sqrt{1 - \bar{\alpha}_t}\,\boldsymbol{\epsilon},
\qquad \boldsymbol{\epsilon} \sim \mathcal{N}(\mathbf{0}, \mathbf{I})
\setlength{\abovedisplayskip}{5pt}
\setlength{\belowdisplayskip}{5pt}
\label{eq:forward-reparam}
\end{equation}
where the marginals at different steps share the same latent noise $\boldsymbol{\epsilon}$ but apply different signal transformations $\mathbf{A}_t$, defining a non-Markovian forward family analogous to the denoising diffusion implicit model (DDIM) construction~\cite{song2021ddim}.
When $\mathbf{A}_t = \mathbf{I}$ for all $t$, all components are treated uniformly and the forward trajectory remains isotropic, which recovers the standard diffusion formulation as a special case. Appendix~\ref{sec:appendix-diffusion-bg} reviews this standard framework and its connection to our generalized formulation.
Specifically, for a length-$S$ sequence $\mathbf{z} = [z_1, z_2, \ldots, z_S]^\top$, we define the second-order differencing operator $\mathbf{D}_2$ entrywise as
\begin{equation}
[\mathbf{D}_2 \mathbf{z}]_s =
\begin{cases}
0, & s = 1 \\
z_s - z_{s-1}, & s = 2 \\
z_s - 2z_{s-1} + z_{s-2}, & s \geq 3
\end{cases}
\label{eq:D2}
\setlength{\abovedisplayskip}{5pt}
\setlength{\belowdisplayskip}{5pt}
\end{equation}
The first position removes the absolute level, the second retains the first-order difference, and the remaining positions encode second-order differences. {We use second-order differencing as the lowest-order operator that removes both level and linear trend in the interior while retaining local change structure.\footnote{Appendix~\ref{sec:appendix-fp} compares first-, second-, and third-order differencing operators.}} The transition matrix is then instantiated as
\begin{equation}
\mathbf{A}_t = (1 - \lambda_t)\,\mathbf{I} + \lambda_t\,\mathbf{D}_2
\label{eq:At}
\setlength{\abovedisplayskip}{5pt}
\setlength{\belowdisplayskip}{5pt}
\end{equation}
where $\lambda_t \in [0, \lambda_{\max}]$ increases monotonically with the diffusion step, $\lambda_0 = 0$ (so that $\mathbf{A}_0 = \mathbf{I}$), and $\lambda_{\max} < 1$. Early steps therefore remain close to the target $\mathbf{x}_0$ itself, while later steps increasingly emphasize differential structure. The effective signal energy at step $t$ scales with $\bar{\alpha}_t \|\mathbf{A}_t \mathbf{x}_0\|^2$ while the noise covariance $(1-\bar{\alpha}_t)\mathbf{I}$ remains isotropic, so $\mathbf{D}_2$'s low-frequency suppression and high-frequency preservation produce a non-uniform per-frequency signal-to-noise ratio (SNR) that shifts the components retained by the noisy intermediate state toward those less reproducible from the observed history. Appendices~\ref{sec:spectral-D2} and~\ref{sec:freq-snr} formalize this frequency-selective profile, and Section~\ref{sec:effect-component} provides empirical validation.
Because $\mathbf{D}_2$ couples each position to its two preceding neighbors, defining the clean target as the future window alone would prevent it from encoding cross-boundary curvature at the start of the forecast. We therefore instantiate $\mathbf{x}_0$ as an extended forecasting segment that includes a short overlap with the observed history
\begin{equation}
\mathbf{x}_0 = [\mathbf{x}_{\text{label}},\,\mathbf{x}_{\text{fut}}]
\label{eq:extended-target}
\setlength{\abovedisplayskip}{5pt}
\setlength{\belowdisplayskip}{5pt}
\end{equation}
where $\mathbf{x}_{\text{label}}$ is a short label window of length $L_l \geq 2$ taken from the end of the observed history, and the extended target has total length $S = L_l + H$. Appendix~\ref{sec:extended-target-theory} provides the formal analysis.
Although $\mathbf{A}_t$ reshapes the intermediate trajectory, the diffusion endpoint is unchanged. As $\bar{\alpha}_t \to 0$ when $t \to T$, the signal term vanishes regardless of $\mathbf{A}_t$, and the terminal distribution approaches a standard Gaussian $\mathcal{N}(\mathbf{0}, \mathbf{I})$, as verified in Appendix~\ref{sec:terminal-convergence}.

\subsection{Conditional Reverse Process}
\label{sec:reverse}
Since the transition matrix $\mathbf{A}_t$ varies with the diffusion step, the standard DDPM posterior $q(\mathbf{x}_{t-1} \mid \mathbf{x}_t, \mathbf{x}_0)$ no longer directly applies to our non-isotropic forward process. We therefore follow the non-Markovian formulation of DDIM~\cite{song2021ddim} and parameterize the reverse dynamics through a denoiser that predicts the clean target in the value domain.
Specifically, given a noisy state $\mathbf{x}_t$ at step $t$ and the observed history $\mathbf{x}_{\text{hist}}$, the denoiser $f_\theta(\mathbf{x}_t, t, \mathbf{x}_{\text{hist}})$ predicts a clean target $\hat{\mathbf{x}}_0$. This value-domain prediction is then used to recover the corresponding noise estimate
\begin{equation}
\hat{\boldsymbol{\epsilon}} =
\frac{\mathbf{x}_t - \sqrt{\bar{\alpha}_t}\,\mathbf{A}_t\,\hat{\mathbf{x}}_0}
{\sqrt{1 - \bar{\alpha}_t}}
\label{eq:eps-hat}
\setlength{\abovedisplayskip}{5pt}
\setlength{\belowdisplayskip}{5pt}
\end{equation}
For a reverse step from $\mathbf{x}_t$ to an earlier state $\mathbf{x}_s$ with $s < t$, we use the update
\begin{equation}
\mathbf{x}_s =
\sqrt{\bar{\alpha}_s}\,\mathbf{A}_s\,\hat{\mathbf{x}}_0
+
\sqrt{1 - \bar{\alpha}_s - \sigma_t^2}\,\hat{\boldsymbol{\epsilon}}
+
\sigma_t\,\boldsymbol{\epsilon}'
\label{eq:ddim-reverse}
\setlength{\abovedisplayskip}{5pt}
\setlength{\belowdisplayskip}{5pt}
\end{equation}
where $\boldsymbol{\epsilon}' \sim \mathcal{N}(\mathbf{0}, \mathbf{I})$ and $\sigma_t$ controls the stochasticity of the reverse transition under $0 \leq \sigma_t^2 \leq 1 - \bar{\alpha}_s$, which we set to $\sigma_t = 0$ in all experiments. Although this update involves step-dependent matrices $\mathbf{A}_t$ and $\mathbf{A}_s$, it remains consistent with the forward marginals. In particular, if the denoiser is perfect so that $\hat{\mathbf{x}}_0 = \mathbf{x}_0$, then \eqref{eq:ddim-reverse} yields
\begin{equation}
\mathbf{x}_s \sim q(\mathbf{x}_s \mid \mathbf{x}_0)
= \mathcal{N}\!\left(\sqrt{\bar{\alpha}_s}\,\mathbf{A}_s\,\mathbf{x}_0,\,(1-\bar{\alpha}_s)\,\mathbf{I}\right)
\setlength{\abovedisplayskip}{5pt}
\setlength{\belowdisplayskip}{5pt}
\end{equation}
for any choice of $\{\mathbf{A}_t\}$, because the $\mathbf{A}_t$ term cancels exactly in the noise residual $\mathbf{x}_t - \sqrt{\bar{\alpha}_t}\,\mathbf{A}_t\,\mathbf{x}_0$. The formal derivation is provided in Appendix~\ref{sec:marginal-consistency}.
The reverse process operates across two representation domains. The denoiser always predicts $\hat{\mathbf{x}}_0$ in the value domain, while the reverse state $\mathbf{x}_s$ remains in the step-dependent representation induced by $\mathbf{A}_s$, so the term $\mathbf{A}_s \hat{\mathbf{x}}_0$ maps the predicted clean target into the correct signal domain at step $s$. As $s$ decreases, $\mathbf{A}_s$ gradually approaches $\mathbf{I}$ and the reverse trajectory returns to the original value representation.

\vspace{-0.2cm}
\subsection{Stage-Adaptive Conditional Denoiser}
\label{sec:denoiser}
Since the forward process progressively shifts the signal preserved by the noisy intermediate state from the target itself toward its differenced structure, the reverse model should receive conditioning information that supplies both the value-domain content from the observed history and cues about the history's temporal evolution. We therefore design a stage-adaptive conditional denoiser $f_\theta(\mathbf{x}_t, t, \mathbf{x}_{\text{hist}})$ that builds its conditioning from two complementary views of $\mathbf{x}_{\text{hist}}$. The value path encodes the history directly, and the difference path encodes its first-order temporal differences,
\begin{equation}
\mathbf{h}_{\text{val}} = \mathrm{MLP}_{\text{val}}(\mathbf{x}_{\text{hist}}),
\qquad
\mathbf{h}_{\text{diff}} = \mathrm{MLP}_{\text{diff}}(\Delta\mathbf{x}_{\text{hist}})
\setlength{\abovedisplayskip}{5pt}
\setlength{\belowdisplayskip}{5pt}
\label{eq:dual-enc}
\end{equation}
where $\Delta\mathbf{x}_{\text{hist}}$ denotes the sequence of consecutive first-order differences of $\mathbf{x}_{\text{hist}}$. The value representation $\mathbf{h}_{\text{val}}$ preserves level, scale, and trend information, while the difference representation $\mathbf{h}_{\text{diff}}$ captures local velocity and short-term fluctuations. Although the forward process uses the second-order operator $\mathbf{D}_2$, first-order differences suffice for the conditioning path because it only needs to convey the local velocity of the history, while the second-order structure is already encoded in the noisy target through $\mathbf{A}_t$.
Since the signal geometry changes with the diffusion step, a fixed combination of these two paths is not appropriate. We therefore introduce a timestep-adaptive gate
\begin{equation}
\mathbf{g} =
\mathrm{sigmoid}\!\left(
\mathrm{MLP}_{\text{gate}}
\!\left(
[\mathbf{t}_{\text{emb}},\,\bar{\mathbf{h}}_{\text{val}}]
\right)
\right),
\qquad
\mathbf{c} =
\mathbf{h}_{\text{val}} + \mathbf{g} \odot \mathbf{h}_{\text{diff}}
\setlength{\abovedisplayskip}{5pt}
\setlength{\belowdisplayskip}{5pt}
\label{eq:gate-cond}
\end{equation}
where $\mathbf{t}_{\text{emb}}$ is a learned diffusion-step embedding, $\bar{\mathbf{h}}_{\text{val}}$ is the mean of $\mathbf{h}_{\text{val}}$, $\mathrm{MLP}_{\text{gate}}$ is a two-layer MLP with SiLU activation, and $\odot$ denotes element-wise multiplication. The gate raises the contribution of $\mathbf{h}_{\text{diff}}$ at diffusion stages where the noisy state retains more differenced structure and suppresses it when the value-domain content alone suffices.
The conditioning feature $\mathbf{c}$ enters the denoiser through two pathways. It serves as global context for the denoising backbone $h_\theta$, a stack of residual MLP layers, and is separately projected into the target sequence space to form a value-level baseline. The full denoiser output is
\begin{equation}
f_\theta(\mathbf{x}_t, t, \mathbf{x}_{\text{hist}})
=
h_\theta(\mathbf{x}_t, t, \mathbf{c})
+
\mathbf{W}_p\,\mathbf{c}
\setlength{\abovedisplayskip}{5pt}
\setlength{\belowdisplayskip}{5pt}
\label{eq:denoiser-decomp}
\end{equation}
where $\mathbf{W}_p$ is a learned linear projection. The second term supplies a value-domain baseline directly from the conditioning, while $h_\theta$ recovers the remaining structure from the noisy input. Alternative conditioning variants are discussed in Appendix~\ref{sec:appendix-variants}.

\subsection{Training and Sampling}
\label{sec:training}

\textbf{Training objective.}
Before diffusion, the extended target $\mathbf{x}_0 = [\mathbf{x}_{\text{label}},\,\mathbf{x}_{\text{fut}}]$ is instance-normalized along the time dimension to obtain $\tilde{\mathbf{x}}_0$~\cite{kim2022revin}. This normalization removes sample-specific level and scale variation, so the diffusion's modeling capacity is not consumed by absorbing per-sample magnitude shifts. The denoiser is trained to reconstruct this normalized target with a position-weighted objective,
\begin{equation}
\mathcal{L}_{\mathrm{recon}} =
\mathbb{E}_{t,\,\tilde{\mathbf{x}}_0,\,\boldsymbol{\epsilon}}
\left[
\sum_{i=1}^{S} w_i
\left(
f_{\theta,i}(\mathbf{x}_t, t, \mathbf{x}_{\text{hist}})
-
\tilde{x}_{0,i}
\right)^{\!2}
\right],
\setlength{\abovedisplayskip}{5pt}
\setlength{\belowdisplayskip}{5pt}
\label{eq:loss-recon}
\end{equation}
where $t \sim \mathrm{Uniform}\{0, \ldots, T-1\}$, $\mathbf{x}_t$ is sampled from the forward marginal defined on $\tilde{\mathbf{x}}_0$, and $w_i = w_{\mathrm{fut}}$ is a fixed hyperparameter ($w_{\mathrm{fut}} \geq 1$) for future positions and $w_i = 1$ for label positions. This asymmetric weighting keeps the optimization focused on the forecasting segment while ensuring the label window remains well-reconstructed to preserve cross-boundary continuity. Appendix~\ref{sec:training-objective} details the full denoising formulation underlying this objective.
To recover the absolute scale removed by normalization, an auxiliary linear head estimates the instance statistics $(\hat{\mu}, \hat{\nu})$ from the conditioning feature $\mathbf{c}$, where $\mu$ and $\nu$ are the per-instance mean and standard deviation of $\mathbf{x}_0$ along the time dimension. The total loss is
\begin{equation}
\mathcal{L}
=
\mathcal{L}_{\mathrm{recon}}
+
(\hat{\mu} - \mu)^2
+
(\hat{\nu} - \nu)^2
\setlength{\abovedisplayskip}{5pt}
\setlength{\belowdisplayskip}{5pt}
\label{eq:loss}
\end{equation}
where $(\mu, \nu)$ are the true instance normalization statistics. Appendix~\ref{sec:hybrid-loss} establishes that this joint objective also controls the unnormalized prediction error. At inference, the predicted statistics map the denoised output from the normalized target space back to the original value scale.

\textbf{Sampling.}
At inference, forecasting starts from $\mathbf{x}_T \sim \mathcal{N}(\mathbf{0}, \mathbf{I})$ and applies the reverse update in \eqref{eq:ddim-reverse} with $\sigma_t = 0$. Since the reverse step is defined for arbitrary step pairs $(t, s)$ and the terminal distribution approaches a standard Gaussian, the standard DDIM sub-sampling strategy applies directly. We therefore use a subset of reverse steps during inference to reduce sampling cost without retraining. After de-normalization using the predicted statistics $(\hat{\mu}, \hat{\nu})$, the first $L_l$ label positions are discarded and the remaining $H$ positions serve as the forecast sample.

\vspace{-0.2cm}
\section{Experiment Settings}
\label{sec:experiment-settings}
\vspace{-0.2cm}
\subsection{Datasets}
\vspace{-0.2cm}
We evaluate on seven widely used time series datasets spanning diverse temporal characteristics, namely \texttt{Electricity}, \texttt{ETTm2}~\cite{zhou2021informer}, \texttt{Exchange Rate}, \texttt{Traffic}, \texttt{Solar-Energy}~\cite{lai2018lstnet}, \texttt{Weather}~\cite{wu2021autoformer}, and \texttt{Wind}~\cite{wang2025matsd}.
Each dataset is split chronologically into training, validation, and test sets following standard forecasting protocol, with a ratio of 6:2:2 for ETTm2 and 7:1:2 for the remaining six datasets.
Following the prior studies \cite{wu2021autoformer, nie2023patchtst}, we use a lookback length $L = 96$ for both \textsc{DiffDiff} and all baselines, and evaluate across four prediction horizons $H \in \{96, 192, 336, 720\}$.
Probabilistic forecasting quality is measured by the continuous ranked probability score (CRPS)~\cite{gneiting2007crps} where lower values indicate better calibrated and sharper forecasts.
\vspace{-0.1cm}
\subsection{Baselines and Comparing Approaches}
We compare \textsc{DiffDiff} against three controlled ablations that independently isolate its forward process and its denoiser.
The \emph{Standard} variant pairs an isotropic Gaussian forward with a plain history-encoded denoiser, removing both design choices.
The \emph{Diff-forward} variant keeps the differencing forward process but reverts the denoiser to a standard conditioning pathway that encodes only the value-domain content of the history, so that any gain over Standard isolates the forward process.
The \emph{Std+G} variant pairs an isotropic Gaussian forward with the stage-adaptive conditional denoiser, so that any gain over Standard isolates the denoiser.
For comparison with existing diffusion-based forecasters, we include isotropic-forward designs CSDI~\cite{tashiro2021csdi} and TimeDiff~\cite{shen2023timediff}, methods that modify the diffusion process from within, namely TMDM~\cite{li2024tmdm}, NsDiff~\cite{ye2025nsdiff}, and MA-TSD~\cite{wang2025matsd}, and D3U~\cite{li2025d3u}, which pre-cleans the residual through an external decomposition before applying diffusion.

\vspace{-0.1cm}
\subsection{Implementation Details}
For \textsc{DiffDiff}, $\lambda_t$ increases linearly from $0$ to $\lambda_{\max}=0.5$ along the diffusion chain, with label window length $L_l=48$ and future-position loss weight $w_{\text{fut}}=5$. The diffusion uses $T=50$ steps under a cosine $\bar{\alpha}_t$ schedule, with inference via 10-step DDIM sub-sampling and 100 stochastic samples. The denoiser is a stack of three residual MLP layers with model dimension 128, feedforward dimension 512, and dropout 0.1. Optimization uses Adam~\cite{kingma2015adam} at learning rate $2\times10^{-4}$, weight decay $10^{-5}$, and batch size 64 for up to 100 epochs, with all reported results averaged over five random seeds.

\vspace{-0.2cm}
\section{Results and Analysis}
\vspace{-0.1cm}
\subsection{Overall Results}
\label{sec:main-results}

\begin{table}[t]
\centering
\caption{Ablation results on seven benchmarks across four prediction horizons. For every column the best value across the four methods is \textbf{bold} and the second-best is \underline{underlined}. Standard uses an isotropic forward with a plain denoiser. The other three rows replace the forward only (Diff-forward), the denoiser only (Std+G), or both (\textsc{DiffDiff}). The rightmost column reports per-row wins.}
\vspace{-0.2cm}
\label{tab:ablation}
\footnotesize
\setlength{\tabcolsep}{2.0pt}
\resizebox{\textwidth}{!}{
\begin{tabular}{c|c| ccc| ccc| ccc| ccc| ccc| ccc| ccc| |c}
\toprule
\multirow{2}{*}{Model} & \multirow{2}{*}{$H$}
 & \multicolumn{3}{c|}{Electricity}
 & \multicolumn{3}{c|}{ETTm2}
 & \multicolumn{3}{c|}{Exchange}
 & \multicolumn{3}{c|}{Traffic}
 & \multicolumn{3}{c|}{Weather}
 & \multicolumn{3}{c|}{Solar}
 & \multicolumn{3}{c||}{Wind}
 & \multirow{2}{*}{Wins} \\
\cmidrule(lr){3-5} \cmidrule(lr){6-8} \cmidrule(lr){9-11} \cmidrule(lr){12-14} \cmidrule(lr){15-17} \cmidrule(lr){18-20} \cmidrule(lr){21-23}
 & & MSE & MAE & CRPS & MSE & MAE & CRPS & MSE & MAE & CRPS & MSE & MAE & CRPS & MSE & MAE & CRPS & MSE & MAE & CRPS & MSE & MAE & CRPS & \\
\midrule
\multirow{4}{*}{\rotatebox{90}{Standard}}
& 96 & .371 & .442 & .193 & .134 & .285 & .122 & .103 & .251 & .110 & .221 & .320 & .149 & .125 & .268 & .116 & .650 & .547 & .248 & .645 & .567 & .246 & 0 \\
& 192 & .403 & .467 & .195 & .153 & .303 & .123 & .210 & .365 & .157 & .208 & .305 & .142 & .188 & .343 & .144 & .642 & .564 & .246 & .892 & .696 & \underline{.301} & 0 \\
& 336 & .951 & .560 & .206 & .207 & .361 & .147 & .656 & .626 & .261 & .215 & .315 & .147 & .256 & .397 & .168 & .515 & .498 & .209 & .980 & .753 & .328 & 0 \\
& 720 & .565 & .561 & .223 & .217 & .368 & .151 & 1.024 & .859 & .407 & .351 & .431 & .180 & .500 & .593 & .254 & 1.362 & .823 & .348 & 1.249 & .890 & .361 & 0 \\
\midrule
\multirow{4}{*}{\rotatebox{90}{Diff-F}}
 & 96 & \underline{.281} & \underline{.369} & \underline{.152} & \bf .067 & \bf .189 & \bf .078 & \bf .098 & \bf .239 & \bf .099 & \underline{.157} & \bf .233 & \bf .107 & \bf .093 & \bf .222 & \underline{.095} & \underline{.490} & \underline{.447} & \underline{.197} & \underline{.637} & .563 & \underline{.234} & 10 \\
 & 192 & \underline{.307} & \underline{.384} & \underline{.159} & \underline{.104} & \underline{.240} & \underline{.099} & \bf .183 & \bf .341 & \bf .141 & \bf .150 & \bf .228 & \bf .104 & \underline{.138} & \underline{.275} & \underline{.115} & \underline{.495} & \underline{.453} & \underline{.192} & \underline{.854} & \underline{.683} & \bf .278 & 7 \\
 & 336 & \bf .344 & \bf .410 & \bf .167 & \underline{.132} & \underline{.278} & \underline{.114} & \bf .330 & \bf .468 & \bf .193 & \underline{.151} & \underline{.233} & \bf .105 & \underline{.217} & \underline{.350} & \underline{.145} & .432 & .421 & .173 & \bf .912 & \underline{.746} & \underline{.305} & 8 \\
 & 720 & \underline{.442} & \underline{.477} & \underline{.197} & \underline{.188} & \underline{.337} & \underline{.138} & \underline{.862} & \underline{.790} & \underline{.358} & .177 & .265 & .121 & .342 & \underline{.444} & \underline{.184} & .492 & \underline{.456} & \underline{.190} & \underline{.979} & .799 & \underline{.323} & 0 \\
\midrule
\multirow{4}{*}{\rotatebox{90}{Std+G}}
 & 96 & .294 & .377 & .164 & .087 & .219 & .089 & .103 & .254 & .112 & .196 & .283 & .126 & \underline{.103} & \underline{.243} & .104 & .566 & .488 & .212 & .649 & \underline{.562} & .249 & 0 \\
 & 192 & .351 & .437 & .179 & .120 & .277 & .118 & .207 & .378 & .157 & .172 & .261 & .126 & .158 & .296 & .125 & .512 & .486 & .196 & .891 & .695 & .302 & 0 \\
 & 336 & .450 & .514 & .200 & .170 & .304 & .127 & \underline{.376} & \underline{.491} & .212 & .179 & .255 & .126 & .223 & .355 & .153 & \underline{.413} & \underline{.410} & \underline{.169} & \underline{.949} & .774 & .324 & 0 \\
 & 720 & .449 & .479 & .199 & \underline{.188} & .339 & .143 & .937 & .805 & .363 & \underline{.171} & \underline{.257} & \underline{.117} & \underline{.341} & .446 & .185 & \underline{.488} & .458 & .193 & .987 & \underline{.797} & .328 & 0 \\
\midrule
\multirow{4}{*}{\rotatebox{90}{\textsc{DiffDiff}}}
 & 96 & \bf .269 & \bf .362 & \bf .150 & \underline{.068} & \underline{.190} & \underline{.079} & \underline{.101} & \underline{.246} & \underline{.103} & \bf .156 & \underline{.234} & \underline{.108} & \bf .093 & \bf .222 & \bf .093 & \bf .463 & \bf .435 & \bf .191 & \bf .599 & \bf .547 & \bf .226 & 13 \\
 & 192 & \bf .299 & \bf .381 & \bf .158 & \bf .102 & \bf .239 & \bf .098 & \underline{.198} & \underline{.358} & \underline{.147} & \underline{.151} & \underline{.231} & \underline{.106} & \bf .135 & \bf .271 & \bf .112 & \bf .483 & \bf .448 & \bf .191 & \bf .815 & \bf .675 & \bf .278 & 15 \\
 & 336 & \underline{.349} & \underline{.413} & \underline{.171} & \bf .128 & \bf .273 & \bf .110 & .388 & .495 & \underline{.207} & \bf .148 & \bf .232 & \underline{.106} & \bf .202 & \bf .332 & \bf .137 & \bf .389 & \bf .395 & \bf .163 & \bf .912 & \bf .742 & \bf .300 & 14 \\
 & 720 & \bf .435 & \bf .474 & \bf .196 & \bf .182 & \bf .330 & \bf .133 & \bf .735 & \bf .728 & \bf .327 & \bf .168 & \bf .252 & \bf .114 & \bf .328 & \bf .434 & \bf .178 & \bf .485 & \bf .455 & \bf .189 & \bf .966 & \bf .785 & \bf .317 & 21 \\
\bottomrule
\end{tabular}
\vspace{-0.2cm}
}
\end{table}

\begin{table}[t]
\centering
        \caption{Forecasting results on seven benchmarks across four prediction horizons. For every column the best value across methods is \textbf{bold} and the second-best is \underline{underlined}. The rightmost column reports per-row wins.}
\label{tab:main}
\vspace{-0.2cm}
\footnotesize
\setlength{\tabcolsep}{2.0pt}
\resizebox{\textwidth}{!}{
\begin{tabular}{c|c| ccc| ccc| ccc| ccc| ccc| ccc| ccc| |c}
\toprule
\multirow{2}{*}{Model} & \multirow{2}{*}{$H$}
 & \multicolumn{3}{c|}{Electricity}
 & \multicolumn{3}{c|}{ETTm2}
 & \multicolumn{3}{c|}{Exchange}
 & \multicolumn{3}{c|}{Traffic}
 & \multicolumn{3}{c|}{Weather}
 & \multicolumn{3}{c|}{Solar}
 & \multicolumn{3}{c||}{Wind}
 & \multirow{2}{*}{Wins} \\
\cmidrule(lr){3-5} \cmidrule(lr){6-8} \cmidrule(lr){9-11} \cmidrule(lr){12-14} \cmidrule(lr){15-17} \cmidrule(lr){18-20} \cmidrule(lr){21-23}
 & & MSE & MAE & CRPS & MSE & MAE & CRPS & MSE & MAE & CRPS & MSE & MAE & CRPS & MSE & MAE & CRPS & MSE & MAE & CRPS & MSE & MAE & CRPS & \\
\midrule
\multirow{4}{*}{\rotatebox{90}{CSDI}}
 & 96 & 1.18 & .853 & .332 & .444 & .535 & .257 & .377 & .505 & .223 & 1.96 & 1.20 & .453 & .479 & .568 & .233 & 1.38 & .987 & .343 & .968 & .786 & .296 & 0 \\
 & 192 & 7.77 & 2.24 & .892 & 2.41 & 1.29 & .539 & .753 & .715 & .277 & 1.99 & 1.21 & .456 & 3.64 & 1.59 & .639 & 2.63 & 1.30 & .553 & 3.50 & 1.52 & .640 & 0 \\
 & 336 & 13.0 & 1.69 & 2.37 & 7.27 & 2.38 & 1.06 & 3.84 & 1.65 & .592 & 1.98 & 1.21 & .457 & 5.24 & 2.05 & .805 & 3.30 & 1.52 & .644 & 9.74 & 2.61 & 1.09 & 0 \\
 & 720 & 1.03 & .814 & .327 & 6.61 & 2.25 & 1.12 & 12.1 & 2.73 & .902 & 2.00 & 1.21 & .457 & 14.0 & 3.42 & 1.57 & 3.52 & 1.48 & .633 & 13.0 & 3.20 & 1.43 & 0 \\
\cmidrule(lr){1-24}
\multirow{4}{*}{\rotatebox{90}{TMDM}}
 & 96 & .315 & .412 & .168 & .072 & .200 & \underline{.081} & \underline{.111} & .255 & \bf .102 & .205 & .272 & .117 & .123 & .257 & .105 & \bf .359 & \bf .346 & \bf .136 & .663 & \underline{.560} & \underline{.231} & 4 \\
 & 192 & .408 & .472 & .191 & .125 & .269 & .109 & .418 & .469 & .200 & .194 & .278 & .116 & .163 & .298 & \underline{.121} & \bf .291 & \bf .322 & \bf .123 & .885 & .698 & \bf .274 & 4 \\
 & 336 & .454 & .495 & .200 & .158 & .306 & .123 & .620 & .598 & .245 & .181 & .257 & \underline{.109} & \underline{.226} & \underline{.348} & \underline{.141} & \bf .291 & \bf .333 & \bf .123 & \underline{.904} & \bf .704 & \bf .284 & 5 \\
 & 720 & .550 & .529 & .215 & .233 & .375 & .150 & \bf .726 & \bf .716 & \bf .312 & .194 & \underline{.267} & .123 & \underline{.329} & .438 & .182 & .496 & .472 & .199 & 1.11 & .809 & .338 & 3 \\
\cmidrule(lr){1-24}
\multirow{4}{*}{\rotatebox{90}{TimeDiff}}
 & 96 & .568 & .579 & .283 & .126 & .273 & .132 & .113 & .257 & .119 & .277 & .362 & .176 & .136 & .273 & .132 & 1.03 & .736 & .362 & .917 & .654 & .323 & 0 \\
 & 192 & .546 & .556 & .267 & .155 & .303 & .147 & .270 & .395 & .193 & .261 & .350 & .170 & .176 & .307 & .146 & 1.46 & .946 & .467 & 1.26 & .810 & .400 & 0 \\
 & 336 & .565 & .570 & .273 & .189 & .337 & .160 & .477 & .529 & .260 & .264 & .366 & .177 & .237 & .355 & .173 & .969 & .772 & .369 & 1.59 & .920 & .441 & 0 \\
 & 720 & .716 & .671 & .322 & .222 & .373 & .182 & 1.30 & .884 & .437 & .303 & .389 & .185 & .352 & .440 & .215 & 1.44 & .938 & .462 & 1.70 & 1.01 & .505 & 0 \\
\cmidrule(lr){1-24}
\multirow{4}{*}{\rotatebox{90}{NsDiff}}
 & 96 & .345 & .431 & .172 & .093 & .230 & .095 & .123 & .270 & .111 & .203 & .292 & \underline{.110} & .101 & .236 & \underline{.101} & \underline{.460} & .493 & \underline{.170} & .664 & .576 & .245 & 0 \\
 & 192 & .355 & .436 & .175 & .139 & .287 & .116 & .317 & .427 & .185 & .187 & .276 & .112 & .158 & .294 & .122 & \underline{.439} & .516 & \underline{.171} & .888 & \underline{.680} & .291 & 0 \\
 & 336 & .461 & .503 & .201 & .184 & .331 & .133 & .404 & \underline{.491} & \underline{.213} & .196 & .297 & \underline{.109} & .230 & .354 & .146 & .548 & .537 & \underline{.161} & 1.09 & .797 & .335 & 0 \\
 & 720 & \bf .404 & \bf .470 & \bf .189 & .255 & .399 & .159 & .863 & \underline{.725} & \underline{.323} & .199 & .290 & \underline{.119} & .334 & \bf .430 & \bf .176 & \bf .440 & \underline{.460} & \bf .179 & 1.13 & \bf .776 & \underline{.329} & 8 \\
\cmidrule(lr){1-24}
\multirow{4}{*}{\rotatebox{90}{MA-TSD}}
 & 96 & \underline{.284} & \underline{.375} & \underline{.157} & \underline{.070} & \underline{.199} & .085 & \bf .101 & \underline{.249} & .112 & \underline{.166} & \underline{.246} & .112 & \underline{.096} & \underline{.230} & .102 & .539 & .499 & .224 & \underline{.614} & .569 & .249 & 1 \\
 & 192 & \underline{.300} & \underline{.383} & \underline{.161} & \underline{.106} & \underline{.246} & .106 & \bf .190 & \bf .354 & \underline{.160} & \underline{.158} & \underline{.239} & \underline{.110} & \underline{.146} & \underline{.289} & .127 & .500 & .456 & .200 & \underline{.848} & .713 & .312 & 2 \\
 & 336 & \bf .347 & \underline{.415} & \underline{.178} & \underline{.134} & \underline{.280} & \underline{.122} & \bf .334 & \bf .479 & .221 & \underline{.156} & \underline{.242} & .113 & \underline{.226} & .361 & .161 & .450 & .421 & .180 & \bf .891 & .752 & .332 & 4 \\
 & 720 & \underline{.432} & \underline{.472} & .205 & \underline{.192} & \underline{.335} & \underline{.145} & .820 & .768 & .370 & \underline{.188} & .278 & .131 & .375 & .477 & .214 & .547 & .475 & .206 & \bf .959 & .788 & .350 & 1 \\
\cmidrule(lr){1-24}
\multirow{4}{*}{\rotatebox{90}{D3U}}
 & 96 & .357 & .433 & .174 & .075 & \underline{.199} & \underline{.081} & .187 & .326 & .155 & .186 & .278 & .132 & .124 & .258 & .106 & 1.09 & .657 & .289 & 1.03 & .733 & .316 & 0 \\
 & 192 & .365 & .436 & .174 & .121 & .259 & \underline{.103} & .687 & .608 & .289 & .210 & .307 & .161 & .206 & .340 & .140 & .995 & .632 & .278 & 1.21 & .821 & .364 & 0 \\
 & 336 & .495 & .506 & .203 & .163 & .307 & .124 & 1.78 & 1.03 & .492 & .199 & .293 & .135 & .263 & .381 & .157 & .604 & .502 & .205 & 1.44 & .918 & .411 & 0 \\
 & 720 & .547 & .551 & .221 & .255 & .387 & .159 & 3.23 & 1.38 & .666 & .245 & .322 & .142 & .439 & .498 & .209 & .717 & .555 & .237 & 1.87 & 1.07 & .492 & 0 \\
\cmidrule(lr){1-24}
\multirow{4}{*}{\rotatebox{90}{\textsc{DiffDiff}}}
 & 96 & \bf .269 & \bf .362 & \bf .150 & \bf .068 & \bf .190 & \bf .079 & \bf .101 & \bf .246 & \underline{.103} & \bf .156 & \bf .234 & \bf .108 & \bf .093 & \bf .222 & \bf .093 & .463 & \underline{.435} & .191 & \bf .599 & \bf .547 & \bf .226 & 17 \\
 & 192 & \bf .299 & \bf .381 & \bf .158 & \bf .102 & \bf .239 & \bf .098 & \underline{.198} & \underline{.358} & \bf .147 & \bf .151 & \bf .231 & \bf .106 & \bf .135 & \bf .271 & \bf .112 & .483 & \underline{.448} & .191 & \bf .815 & \bf .675 & \underline{.278} & 15 \\
 & 336 & \underline{.349} & \bf .413 & \bf .171 & \bf .128 & \bf .273 & \bf .110 & \underline{.388} & .495 & \bf .207 & \bf .148 & \bf .232 & \bf .106 & \bf .202 & \bf .332 & \bf .137 & \underline{.389} & \underline{.395} & .163 & .912 & \underline{.742} & \underline{.300} & 12 \\
 & 720 & .435 & .474 & \underline{.196} & \bf .182 & \bf .330 & \bf .133 & \underline{.735} & .728 & .327 & \bf .168 & \bf .252 & \bf .114 & \bf .328 & \underline{.434} & \underline{.178} & \underline{.485} & \bf .455 & \underline{.189} & \underline{.966} & \underline{.785} & \bf .317 & 9 \\
\bottomrule
\end{tabular}
\vspace{-0.2cm}
}
\end{table}

In Table~\ref{tab:ablation}, Standard wins no setting while the two differencing-forward variants account for every win, identifying the forward process as the primary source of improvement.
Std+G improves over Standard on most settings yet wins no cell in the four-way comparison, confirming that the denoiser carries genuine modeling power yet cannot substitute for the reshaped corruption path.
At $H{=}720$, where long-range reconstruction is hardest, \textsc{DiffDiff} wins every setting while Diff-forward wins none, leaving the two designs complementary rather than redundant.

In Table~\ref{tab:main}, \textsc{DiffDiff} wins 53 of 84 settings ahead of six diffusion baselines.
CSDI, originally an imputation model, deteriorates sharply at long horizons since conditioning on observed points fails to extrapolate over long gaps.
TimeDiff combines a history-conditioned UNet with an isotropic forward, showing that denoiser-side conditioning alone cannot compensate for a task-agnostic corruption path.
NsDiff concentrates all 8 of its wins at $H{=}720$, where its learned non-stationary variance schedule meets prediction windows long enough to include regime changes.
TMDM is the strongest baseline with 16 wins, concentrated on Solar at $H \leq 336$ and Exchange at $H{=}720$, where a persistent regime mean leaves the differencing operator little additional structure to exploit.
D3U obtains no first-place result, with degradation sharpest on regime-shifting series at long horizons where its externally trained point predictor's errors enter the residual the diffusion stage models and the deterministic-stochastic boundary cannot adjust once frozen. Internalizing the asymmetry along the trajectory removes this dependence.
MA-TSD's wins concentrate on Exchange and Wind, whereas \textsc{DiffDiff}'s largest margins appear on Electricity, ETTm2, Traffic, and Weather, where forecast windows place substantial energy in components the history under-determines.

%

%
\begin{figure}[t]
\centering
\includegraphics[width=\linewidth, trim=0 12 0 0]{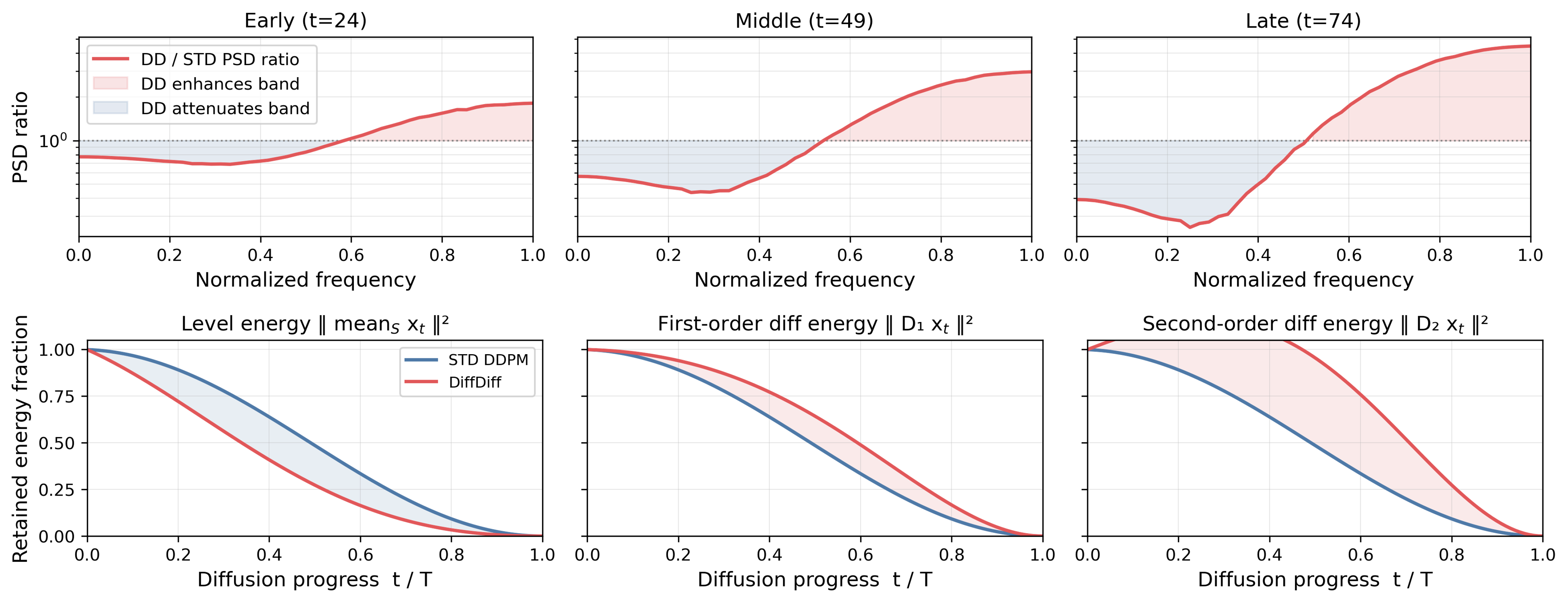}
\caption{Reshaping of the corruption trajectory by the differencing forward process. \textbf{Top}: Per-frequency power spectral density (PSD) ratio of \textsc{DiffDiff} to the standard isotropic forward at three diffusion progresses on \texttt{electricity}, with values above (below) unity marking enhanced (attenuated) bands. \textbf{Bottom}: Retained energy fraction of the window mean and the first- and second-order increments under each forward path along the diffusion progress on \texttt{solar}.}
\label{fig:effect-forward}
\vspace{-0.2cm}
\end{figure}

%
\begin{figure}[t]
\centering
\includegraphics[width=\linewidth, trim=0 20 0 0]{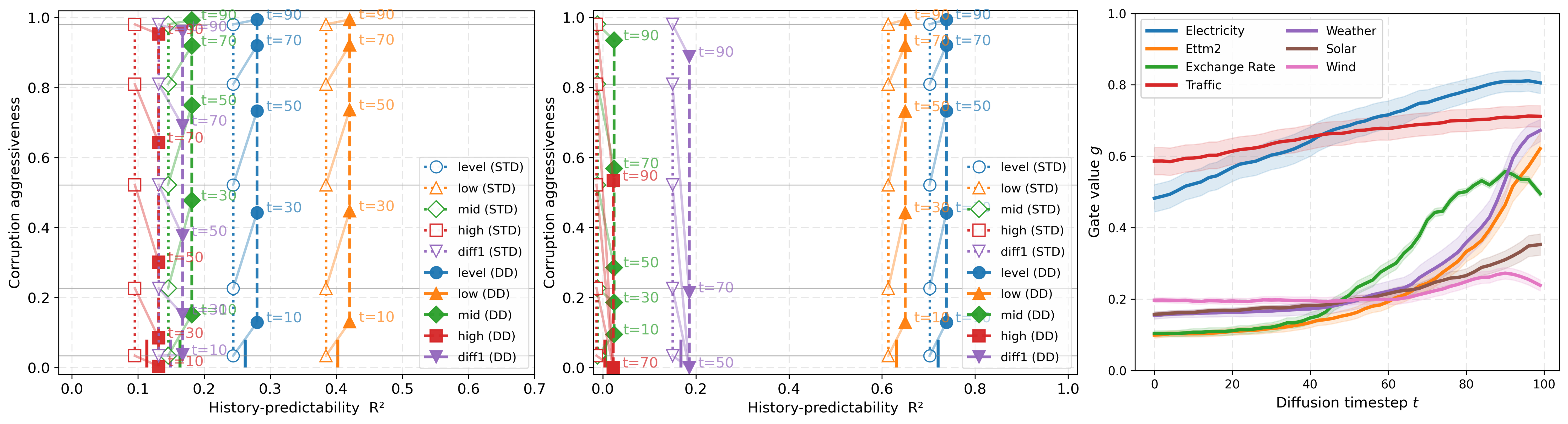}
\caption{Component-level corruption alignment and gate dynamics. \textbf{Left:} Per-component coefficient of determination $R^2$ (horizontal) versus corruption aggressiveness (vertical) on \texttt{wind} and \texttt{exchange-rate}, with each component reducing a window to a scalar, namely the window mean (level), summed fast Fourier transform (FFT) magnitudes in three frequency bands (low, mid, high), and the mean first-order absolute difference (diff1). Components are traced across five diffusion progresses, with open markers for the standard forward and filled markers for \textsc{DiffDiff}, jittered horizontally for visibility. \textbf{Right:} Gate activation of the differential branch versus diffusion timestep across seven datasets, with shaded bands showing $\pm 1$ standard deviation.}
\label{fig:effect-component-gate}
\vspace{-0.3cm}
\end{figure}

%
%
\vspace{-0.1cm}
\subsection{Spectral Reshaping by the Differencing Forward Process}
\label{sec:effect-forward}

As shown in Figure~\ref{fig:effect-forward}, we strip the additive noise from both forwards and compare what each path retains at three diffusion progresses, in the frequency domain (top) and across three operator-aligned components (bottom).
The reshape unfolds progressively along the chain rather than as a single fixed transform, with the PSD ratio sitting near unity at $t/T{\approx}0.25$ and carving a sharp split between attenuated low-frequency and amplified high-frequency content by $t/T{\approx}0.75$.
{Because both paths share the same noise schedule, this split reflects a change in the signal representation presented to the denoiser at a matched noise level.}
In the operator domain, \textsc{DiffDiff} attenuates the level energy faster than the standard forward, preserves the first-order increment longer, and preserves the second-order increment most of all, identifying $\mathbf{D}_2$ as a structural filter aligned with the order of differencing rather than tuned to arbitrary frequency bands.
{The retained-energy curves separate at intermediate stages, where differential content remains visible after much of the level information has decayed.}
With isotropic noise and selectively attenuated signal, the SNR redistributes toward high-frequency content and away from low-frequency content, turning the corruption trajectory from a uniform attenuator into a frequency-selective preserver.

%
%
\vspace{-0.2cm}
\subsection{Predictability-Driven Corruption Reallocation}
\label{sec:effect-component}

To check whether the corruption budget tracks history-availability, we trace each signal component along the diffusion chain in an $(R^2, \text{aggressiveness})$ plane, with $R^2$ measuring the linear-regression fit from the matched component in the history window to that in the future window and aggressiveness equal to $1 - \bar{\alpha}_t \|\phi(\mathbf{A}_t \mathbf{x}_0)\|^2 / \|\phi(\mathbf{x}_0)\|^2$ at diffusion progress $t/T$, capturing both operator-induced and noise-schedule-induced energy decay.
Wind concentrates all five components into a narrow $R^2$ band while exchange rate spreads them broadly, and on both the predictability ranking lines up with the spectral one, making selectivity along frequency equivalent to selectivity along predictability.
{The contrast is strongest in the level and low-frequency components, whose higher $R^2$ on Exchange Rate indicates stronger linear predictability from the observed history.}
Under the standard forward the five components climb in lockstep regardless of $R^2$, whereas \textsc{DiffDiff} breaks this into a fan-shaped spread, with high-$R^2$ components rising above the standard reference and low-$R^2$ ones falling below.
{As shown in the left panels of Figure~\ref{fig:effect-component-gate}, corruption aggressiveness is positively associated with component predictability in both datasets despite their different $R^2$ ranges.}
At $t/T{=}0.5$ the component-wise correlation between $R^2$ and aggressiveness rises from zero under the standard forward to $+0.76$ on wind and $+0.58$ on exchange rate under \textsc{DiffDiff}, with more history-predictable components attenuated more heavily and less predictable ones preserved.
The forward process therefore spends its attenuation budget on the components the history can already supply, freeing the diffusion's representational capacity for the harder, history-under-determined ones.

%
%
\vspace{-0.2cm}
\subsection{Stage-Adaptive Gating of the Differential Branch}
\label{sec:effect-gate}

To examine the gate's adaptation to timestep and dataset, we sweep the diffusion timestep on the trained model and record the mean gate value over batch samples and hidden dimensions at each step.
As shown in the right panel of Figure~\ref{fig:effect-component-gate}, the gate is far from constant, and the seven datasets separate into three behavioral regimes.
Electricity and Traffic, both with strong periodic structure, hold the gate above $0.5$ from the start and lift it further with $t$, keeping the differential branch open throughout the chain.
ETTm2, Exchange Rate, and Weather hold the gate near its initialization in early steps and admit it sharply in the late low-SNR portion, drawing on differential evidence only as the value signal degrades.
{Their late activation suggests that differential history becomes more useful as the target representation shifts toward local changes and direct value cues in the noisy target become less available.}
Solar and Wind suppress the gate throughout the chain, indicating that for these highly stochastic series the differential branch extracts mostly noise rather than usable structure.
{Because the value path remains active for every gate value, suppressing the differential branch still preserves access to the history's level and trend.}
The gate's stage- and dataset-adaptive admission complements the forward operator's predictability-aligned suppression by re-supplying history-anchored content when the noisy target no longer carries it.

\begin{figure}[t]
\centering
\includegraphics[width=0.95\textwidth]{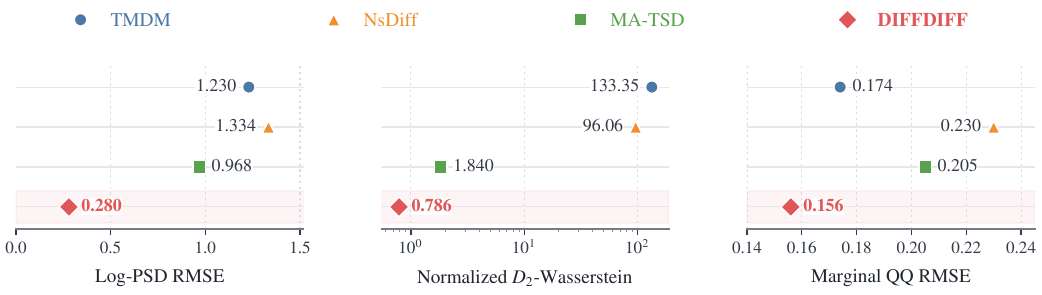}
\caption{{Spectral, local-change, and marginal fidelity of generated forecasts. \textbf{Left:} root mean square error (RMSE) between log power spectral densities (PSDs). \textbf{Middle:} normalized second-order-difference Wasserstein distance on a logarithmic horizontal axis. \textbf{Right:} marginal quantile--quantile (QQ) RMSE. Lower is better for all three metrics. Values are medians computed over all seven datasets and four prediction horizons.}}
\label{fig:distribution-fidelity}
\vspace{-0.2cm}
\end{figure}
%
\vspace{-0.2cm}
\subsection{{Distributional Fidelity of Generated Forecasts}}
\label{sec:distribution-fidelity}

{To further examine whether reallocating modeling capacity improves the structural fidelity of generated trajectories, we evaluate complementary diagnostics of spectral structure, local changes, and marginal distributions. As shown in Figure~\ref{fig:distribution-fidelity}, we compare \textsc{DiffDiff} with TMDM, NsDiff, and MA-TSD using log power spectral density RMSE, normalized second-order-difference Wasserstein distance, and marginal quantile--quantile RMSE.\footnote{Complete metric definitions are provided in Appendix~\ref{sec:appendix-distribution-fidelity}.} Unlike point-forecast MSE and MAE and position-wise CRPS, these diagnostics characterize how predicted values are distributed and organized within trajectories. \textsc{DiffDiff} achieves the lowest median error on all three diagnostics, with larger advantages in spectral and local-change fidelity than in marginal QQ RMSE. This contrast indicates that the main improvement lies in organizing plausible values into realistic temporal dynamics rather than merely adjusting their marginal distribution. In contrast, TMDM and NsDiff retain marginal QQ errors on the same scale as \textsc{DiffDiff} but exhibit normalized second-order-difference Wasserstein distances more than two orders of magnitude larger, showing that plausible pointwise amplitudes do not ensure realistic local temporal dependence. Moreover, \textsc{DiffDiff} more closely matches both the global allocation of variation across frequencies and the distribution of local curvature across generated trajectories, consistent with the differencing forward process retaining target-derived dynamic structure while the value-domain conditioning path anchors level and trend.}

%
\vspace{-0.2cm}
\subsection{Case Study}
\label{sec:case-study}

\begin{figure}[t]
\centering
\includegraphics[width=\textwidth, trim=0 10 0 0]{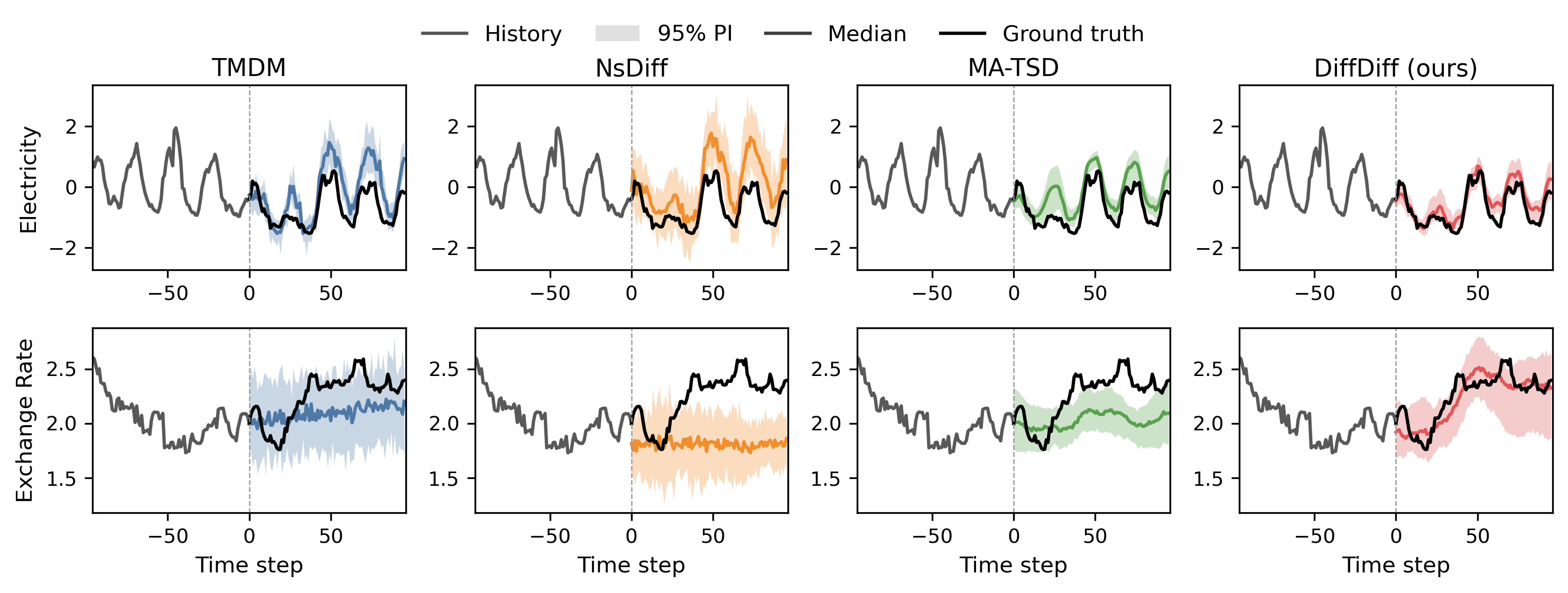}
\vspace{-4mm}
\caption{Per-window $95\%$ prediction interval of \textsc{DiffDiff} and three diffusion baselines on illustrative windows from electricity (top) and exchange rate (bottom) at $H{=}96$, each selected as a top-CRPS-improvement window of \textsc{DiffDiff} over the strongest baseline on that dataset. The gray line shows the history, the black line the ground truth, the colored line the median forecast, and the shaded band the $95\%$ prediction interval.}
\label{fig:case-study}
\vspace{-0.3cm}
\end{figure}

As shown in Figure~\ref{fig:case-study}, we visualize the $95\%$ prediction interval of \textsc{DiffDiff} alongside the three closest diffusion baselines on an illustrative electricity window (top, periodic) and an exchange-rate window (bottom, regime-shifting).
TMDM's bands hedge widely over the latter half of each horizon, with the median tracking the cycle but the spread leaving considerable residual ambiguity.
NsDiff produces visibly jagged predictions whose median and band fluctuate between adjacent time steps, a symptom of its per-cell variance design.
MA-TSD narrows the bands most aggressively on the exchange-rate window, yet the median misses the trend reversal there and overshoots the periodic peaks and troughs on electricity.
{The two windows highlight different forecasting demands, with repeated peaks and troughs requiring phase-sensitive continuation and the exchange-rate reversal requiring adaptation beyond recent persistence.}
{\textsc{DiffDiff}'s median tracks both the periodic cycle and the trend reversal while its prediction intervals remain concentrated around the realized trajectories.}

\vspace{-0.2cm}
\section{Conclusion}
\label{sec:conclusion}
\vspace{-0.1cm}
We show that probabilistic time series forecasting benefits from encoding predictability asymmetry directly into the diffusion trajectory.
\textsc{DiffDiff} progressively transforms the forward state from the target itself toward its second-order differenced structure, while a stage-adaptive conditioning pathway combines value-domain and differential history cues.
This design reallocates model capacity from reconstructing history-anchored content to recovering history-under-determined dynamics.
{Across seven benchmarks, \textsc{DiffDiff} outperforms six diffusion baselines. Analyses of its components and generated distributions confirm that the operator and gate adapt to history predictability along the chain and show that the forecasts preserve the distributional structure of target trajectories.}

\section*{Acknowledgements}

This work is supported by National Key Research and Development Program of China (No. 2023YFC3303802).

\bibliographystyle{plain}
\bibliography{main}

\clearpage
\appendix

\section{Diffusion Model Background}
\label{sec:appendix-diffusion-bg}

This appendix reviews the standard denoising diffusion framework that underpins the \textsc{DiffDiff} design. We follow the formulations in prior studies \cite{ho2020denoising, song2021ddim}, and conclude by showing how the standard framework specializes to a particular case of the generalized forward process introduced in Section~\ref{sec:forward}.

\subsection{Forward Process and Reparameterization}
\label{sec:bg-forward}

A denoising diffusion probabilistic model~\cite{sohl2015deep,ho2020denoising} defines a forward Markov chain that progressively adds Gaussian noise to a data sample $\mathbf{x}_0 \sim q(\mathbf{x}_0)$. The one-step transition is
\begin{equation}
q(\mathbf{x}_t \mid \mathbf{x}_{t-1})
= \mathcal{N}\!\left(\mathbf{x}_t;\;\sqrt{\alpha_t}\,\mathbf{x}_{t-1},\;\beta_t\,\mathbf{I}\right),
\label{eq:bg-one-step}
\end{equation}
where $\beta_t \in (0,1)$ is the noise variance at step $t$ and $\alpha_t = 1 - \beta_t$. By the reparameterization trick,
\begin{equation}
\mathbf{x}_t = \sqrt{\alpha_t}\,\mathbf{x}_{t-1} + \sqrt{\beta_t}\,\boldsymbol{\epsilon}_{t-1},
\qquad \boldsymbol{\epsilon}_{t-1} \sim \mathcal{N}(\mathbf{0},\mathbf{I}).
\label{eq:bg-reparam-one}
\end{equation}
Substituting $\mathbf{x}_{t-1} = \sqrt{\alpha_{t-1}}\,\mathbf{x}_{t-2} + \sqrt{\beta_{t-1}}\,\boldsymbol{\epsilon}_{t-2}$ and using the fact that the sum of two independent Gaussian noise terms $\mathcal{N}(\mathbf{0},\sigma_1^2\mathbf{I}) + \mathcal{N}(\mathbf{0},\sigma_2^2\mathbf{I}) = \mathcal{N}(\mathbf{0},(\sigma_1^2+\sigma_2^2)\mathbf{I})$, we obtain
\begin{equation}
\mathbf{x}_t = \sqrt{\alpha_t\alpha_{t-1}}\,\mathbf{x}_{t-2}
+ \sqrt{1 - \alpha_t\alpha_{t-1}}\,\boldsymbol{\epsilon},
\qquad \boldsymbol{\epsilon} \sim \mathcal{N}(\mathbf{0},\mathbf{I}),
\end{equation}
since $\alpha_t\beta_{t-1} + \beta_t = \alpha_t(1 - \alpha_{t-1}) + (1 - \alpha_t) = 1 - \alpha_t\alpha_{t-1}$. Applying this recursion from step $t$ down to step~$0$ and defining $\bar{\alpha}_t = \prod_{s=1}^{t}\alpha_s$ gives the closed-form forward marginal
\begin{equation}
q(\mathbf{x}_t \mid \mathbf{x}_0)
= \mathcal{N}\!\left(\mathbf{x}_t;\;\sqrt{\bar{\alpha}_t}\,\mathbf{x}_0,\;(1-\bar{\alpha}_t)\,\mathbf{I}\right),
\label{eq:bg-marginal}
\end{equation}
equivalently $\mathbf{x}_t = \sqrt{\bar{\alpha}_t}\,\mathbf{x}_0 + \sqrt{1-\bar{\alpha}_t}\,\boldsymbol{\epsilon}$ with $\boldsymbol{\epsilon} \sim \mathcal{N}(\mathbf{0},\mathbf{I})$.
This reparameterization allows sampling any noisy state $\mathbf{x}_t$ directly from $\mathbf{x}_0$ without iterating through intermediate steps, and it forms the foundation for both training and the non-Markovian extensions discussed below.

\subsection{Evidence Lower Bound}
\label{sec:bg-elbo}

The reverse process is parameterized as
\begin{equation}
p_\theta(\mathbf{x}_{0:T})
= p(\mathbf{x}_T)\prod_{t=1}^{T} p_\theta(\mathbf{x}_{t-1}\mid\mathbf{x}_t),
\qquad
p_\theta(\mathbf{x}_{t-1}\mid\mathbf{x}_t)
= \mathcal{N}\!\left(\mathbf{x}_{t-1};\;\boldsymbol{\mu}_\theta(\mathbf{x}_t,t),\;\tilde{\beta}_t\,\mathbf{I}\right),
\end{equation}
where $p(\mathbf{x}_T) = \mathcal{N}(\mathbf{0},\mathbf{I})$. Applying Jensen's inequality to the log-marginal likelihood yields the evidence lower bound (ELBO):
\begin{equation}
\log p_\theta(\mathbf{x}_0)
\;\geq\;
\mathbb{E}_{q(\mathbf{x}_{1:T}\mid\mathbf{x}_0)}
\!\left[\log p_\theta(\mathbf{x}_{0:T}) - \log q(\mathbf{x}_{1:T}\mid\mathbf{x}_0)\right].
\end{equation}
Using the Markov factorizations of $q$ and $p_\theta$, the ELBO decomposes into~\cite{ho2020denoising}
\begin{equation}
\begin{aligned}
\mathrm{ELBO}
=& {\mathbb{E}_q\!\left[\log p_\theta(\mathbf{x}_0\mid\mathbf{x}_1)\right]}
\;-\; {D_{\mathrm{KL}}\!\left(q(\mathbf{x}_T\mid\mathbf{x}_0)\,\|\,p(\mathbf{x}_T)\right)}
\\&\;-\; \sum_{t=2}^{T}{D_{\mathrm{KL}}\!\left(q(\mathbf{x}_{t-1}\mid\mathbf{x}_t,\mathbf{x}_0)\,\|\,p_\theta(\mathbf{x}_{t-1}\mid\mathbf{x}_t)\right)}.
\end{aligned}
\label{eq:bg-elbo-decomp}
\end{equation}
The forward posterior $q(\mathbf{x}_{t-1}\mid\mathbf{x}_t,\mathbf{x}_0)$ is Gaussian with
\begin{equation}
q(\mathbf{x}_{t-1}\mid\mathbf{x}_t,\mathbf{x}_0)
= \mathcal{N}\!\left(\mathbf{x}_{t-1};\;\tilde{\boldsymbol{\mu}}_t(\mathbf{x}_t,\mathbf{x}_0),\;\tilde{\beta}_t\,\mathbf{I}\right),
\label{eq:bg-posterior}
\end{equation}
where
\begin{equation}
\tilde{\boldsymbol{\mu}}_t
= \frac{\sqrt{\bar{\alpha}_{t-1}}\,\beta_t}{1-\bar{\alpha}_t}\,\mathbf{x}_0
+ \frac{\sqrt{\alpha_t}\,(1-\bar{\alpha}_{t-1})}{1-\bar{\alpha}_t}\,\mathbf{x}_t,
\qquad
\tilde{\beta}_t
= \frac{(1-\bar{\alpha}_{t-1})}{(1-\bar{\alpha}_t)}\,\beta_t.
\label{eq:bg-posterior-params}
\end{equation}
Since $q(\mathbf{x}_{t-1}\mid\mathbf{x}_t,\mathbf{x}_0)$ and $p_\theta(\mathbf{x}_{t-1}\mid\mathbf{x}_t)$ share the same variance $\tilde{\beta}_t$, each KL term reduces to a squared difference between their means:
\begin{equation}
D_{\mathrm{KL}}\!\left(q(\mathbf{x}_{t-1}\mid\mathbf{x}_t,\mathbf{x}_0)\,\|\,p_\theta(\mathbf{x}_{t-1}\mid\mathbf{x}_t)\right)
= \frac{1}{2\tilde{\beta}_t}
\left\|\tilde{\boldsymbol{\mu}}_t - \boldsymbol{\mu}_\theta(\mathbf{x}_t,t)\right\|^2.
\label{eq:bg-kl-gaussian}
\end{equation}

\subsection{Prediction Parameterizations}
\label{sec:bg-param}

Two equivalent network parameterizations are commonly used.

\paragraph{$\boldsymbol{\epsilon}$-prediction.}
Substituting $\mathbf{x}_0 = (\mathbf{x}_t - \sqrt{1-\bar{\alpha}_t}\,\boldsymbol{\epsilon})/\sqrt{\bar{\alpha}_t}$ into \eqref{eq:bg-posterior-params} and letting $\boldsymbol{\mu}_\theta$ predict $\boldsymbol{\epsilon}$ via a network $\boldsymbol{\epsilon}_\theta(\mathbf{x}_t,t)$, each KL term becomes proportional to $\|\boldsymbol{\epsilon} - \boldsymbol{\epsilon}_\theta(\mathbf{x}_t,t)\|^2$. Dropping the timestep-dependent coefficient yields the simplified training objective \cite{ho2020denoising}:
\begin{equation}
\mathcal{L}_{\mathrm{simple}}
= \mathbb{E}_{t,\,\mathbf{x}_0,\,\boldsymbol{\epsilon}}
\!\left[\left\|\boldsymbol{\epsilon} - \boldsymbol{\epsilon}_\theta(\mathbf{x}_t,t)\right\|^2\right],
\qquad t \sim \mathrm{Uniform}\{1,\ldots,T\}.
\label{eq:bg-eps-loss}
\end{equation}

\paragraph{$\mathbf{x}_0$-prediction.}
Alternatively, a network $f_\theta(\mathbf{x}_t,t)$ can directly predict the clean target $\mathbf{x}_0$. The two parameterizations are linked by the bijection
\begin{equation}
\boldsymbol{\epsilon}
= \frac{\mathbf{x}_t - \sqrt{\bar{\alpha}_t}\,\mathbf{x}_0}{\sqrt{1-\bar{\alpha}_t}},
\qquad
\mathbf{x}_0
= \frac{\mathbf{x}_t - \sqrt{1-\bar{\alpha}_t}\,\boldsymbol{\epsilon}}{\sqrt{\bar{\alpha}_t}}.
\label{eq:bg-bijection}
\end{equation}
Given $\mathbf{x}_t$ and $t$, predicting $\mathbf{x}_0$ is equivalent to predicting $\boldsymbol{\epsilon}$, and the corresponding simplified objective takes the form
\begin{equation}
\mathcal{L}_{{\mathbf{x}_0}}
= \mathbb{E}_{t,\,\mathbf{x}_0,\,\boldsymbol{\epsilon}}
\!\left[\left\|\mathbf{x}_0 - f_\theta(\mathbf{x}_t,t)\right\|^2\right].
\label{eq:bg-x0-loss}
\end{equation}

\subsection{DDIM: Non-Markovian Reverse Process}
\label{sec:bg-ddim}

Prior study \cite{song2021ddim} observe that the training objective in \eqref{eq:bg-eps-loss} depends only on the marginals $q(\mathbf{x}_t\mid\mathbf{x}_0)$ and not on the full joint $q(\mathbf{x}_{1:T}\mid\mathbf{x}_0)$. This observation permits defining a family of non-Markovian reverse transitions that share the same marginals. Specifically, a reverse transition from step $t$ to step $t{-}1$ is defined as
\begin{equation}
q_\sigma(\mathbf{x}_{t-1}\mid\mathbf{x}_t,\mathbf{x}_0)
= \mathcal{N}\!\left(
\mathbf{x}_{t-1};\;
\sqrt{\bar{\alpha}_{t-1}}\,\mathbf{x}_0
+ \sqrt{1-\bar{\alpha}_{t-1}-\sigma_t^2}\cdot
\frac{\mathbf{x}_t - \sqrt{\bar{\alpha}_t}\,\mathbf{x}_0}{\sqrt{1-\bar{\alpha}_t}},\;
\sigma_t^2\,\mathbf{I}
\right).
\label{eq:bg-ddim-reverse}
\end{equation}
To verify marginal consistency, substitute $\mathbf{x}_t = \sqrt{\bar{\alpha}_t}\,\mathbf{x}_0 + \sqrt{1-\bar{\alpha}_t}\,\boldsymbol{\epsilon}$ with $\boldsymbol{\epsilon}\sim\mathcal{N}(\mathbf{0},\mathbf{I})$ into \eqref{eq:bg-ddim-reverse}:
\begin{equation}
\mathbf{x}_{t-1}
= \sqrt{\bar{\alpha}_{t-1}}\,\mathbf{x}_0
+ \sqrt{1-\bar{\alpha}_{t-1}-\sigma_t^2}\,\boldsymbol{\epsilon}
+ \sigma_t\,\boldsymbol{\epsilon}',
\qquad \boldsymbol{\epsilon}' \sim \mathcal{N}(\mathbf{0},\mathbf{I}).
\end{equation}
Marginalizing over $\boldsymbol{\epsilon}$ and $\boldsymbol{\epsilon}'$ gives mean $\sqrt{\bar{\alpha}_{t-1}}\,\mathbf{x}_0$ and variance $(1-\bar{\alpha}_{t-1}-\sigma_t^2)+\sigma_t^2 = 1-\bar{\alpha}_{t-1}$, confirming that $q(\mathbf{x}_{t-1}\mid\mathbf{x}_0) = \mathcal{N}(\sqrt{\bar{\alpha}_{t-1}}\,\mathbf{x}_0,\,(1-\bar{\alpha}_{t-1})\mathbf{I})$ is preserved.

Setting $\sigma_t = 0$ yields a deterministic reverse step:
\begin{equation}
\mathbf{x}_{t-1}
= \sqrt{\bar{\alpha}_{t-1}}\,\mathbf{x}_0
+ \sqrt{\frac{1-\bar{\alpha}_{t-1}}{1-\bar{\alpha}_t}}
\left(\mathbf{x}_t - \sqrt{\bar{\alpha}_t}\,\mathbf{x}_0\right).
\label{eq:bg-ddim-det}
\end{equation}
This extends to arbitrary step pairs $(t,s)$ with $s < t$, enabling accelerated sampling via a subset of reverse steps without retraining.

\subsection{Connection to \textsc{DiffDiff}}
\label{sec:bg-diffdiff}

The standard forward marginal in \eqref{eq:bg-marginal} is a special case of the generalized marginal in \eqref{eq:forward-marginal} with $\mathbf{A}_t = \mathbf{I}$ for all $t$. \textsc{DiffDiff} introduces a step-dependent transition matrix $\mathbf{A}_t = (1-\lambda_t)\mathbf{I} + \lambda_t\mathbf{D}_2$, so the signal mean becomes $\sqrt{\bar{\alpha}_t}\,\mathbf{A}_t\mathbf{x}_0$ while the noise covariance remains isotropic $(1-\bar{\alpha}_t)\mathbf{I}$.
The DDIM reverse framework extends naturally to this setting. Given a denoiser $f_\theta$ that predicts $\hat{\mathbf{x}}_0$ in the value domain, the noise estimate and reverse step become \eqref{eq:eps-hat} and \eqref{eq:ddim-reverse} in the main text. The signal term uses $\mathbf{A}_s$ rather than $\mathbf{A}_t$ to ensure the reverse state lies in the correct representation at step $s$. Marginal consistency of this generalized reverse process is proved in Appendix~\ref{sec:marginal-consistency}.
A practical consequence of introducing $\mathbf{A}_t$ is that $\boldsymbol{\epsilon}$-prediction is no longer interchangeable with $\mathbf{x}_0$-prediction. Under $\boldsymbol{\epsilon}$-prediction, the noise-matching loss $\|\boldsymbol{\epsilon} - \hat{\boldsymbol{\epsilon}}\|^2$ is equivalent to the weighted reconstruction error $\frac{\bar{\alpha}_t}{1-\bar{\alpha}_t}\|\mathbf{A}_t(\mathbf{x}_0 - \hat{\mathbf{x}}_0)\|^2$ (see Appendix~\ref{sec:training-objective}), with the quadratic form $\mathbf{A}_t^\top\mathbf{A}_t$ underpenalizing low-frequency reconstruction error and overpenalizing high-frequency error (Corollary 1). Since the forward operator $\mathbf{A}_t$ already encodes the predictability asymmetry by reshaping the noisy intermediate states, this loss-side reweighting would compound the inductive bias rather than complement it, while standard evaluation metrics weight all frequencies uniformly in the value domain. \textsc{DiffDiff} therefore confines the frequency selectivity to the forward process and adopts the $\mathbf{x}_0$-prediction parameterization, which directly minimizes the value-domain error and stays aligned with the downstream evaluation.

\section{Theoretical Analysis}
\label{sec:appendix-theory}

This appendix provides theoretical analysis of the \textsc{DiffDiff} framework. We establish the spectral properties of the differencing operator (Section~\ref{sec:spectral-D2}), prove marginal consistency of the generalized reverse process (Section~\ref{sec:marginal-consistency}), motivate the training objective from denoising consistency and task alignment (Section~\ref{sec:training-objective}), show the hybrid loss is an upper bound on the unnormalized reconstruction error (Section~\ref{sec:hybrid-loss}), analyze the frequency-dependent signal-to-noise ratio (Section~\ref{sec:freq-snr}), verify terminal distribution convergence (Section~\ref{sec:terminal-convergence}), and formalize the role of the label overlap in preserving cross-boundary information (Section~\ref{sec:extended-target-theory}). All analyses are stated in terms of a single length-$S$ sequence and apply per-channel; for multivariate inputs the same operator and bounds apply independently to each channel.

\subsection{Spectral Properties of the Differencing Operator}
\label{sec:spectral-D2}

\paragraph{Proposition 1 (Spectral characterization of $\mathbf{D}_2$).}
Let $\mathbf{D}_2 \in \mathbb{R}^{S \times S}$ be the second-order differencing operator defined in \eqref{eq:D2}. For a discrete complex exponential $\mathbf{e}_\omega = [1, e^{i\omega}, \ldots, e^{i(S-1)\omega}]^\top$ with $\omega \in [0, \pi]$, the per-element gain at any internal position $s \geq 3$ satisfies
\begin{equation}
\left|[\mathbf{D}_2\,\mathbf{e}_\omega]_s\right|^2
= \left|1 - e^{-i\omega}\right|^4
= (2 - 2\cos\omega)^2.
\label{eq:D2-gain}
\end{equation}

\paragraph{Proof.}
Let $z_s = e^{i(s-1)\omega}$ denote the $s$-th entry of $\mathbf{e}_\omega$ under 1-based indexing. For $s \geq 3$:
\begin{align}
[\mathbf{D}_2\,\mathbf{e}_\omega]_s
&= z_s - 2z_{s-1} + z_{s-2} \notag\\
&= e^{i(s-1)\omega} - 2e^{i(s-2)\omega} + e^{i(s-3)\omega} \notag\\
&= e^{i(s-3)\omega}\!\left(e^{2i\omega} - 2e^{i\omega} + 1\right) \notag\\
&= e^{i(s-3)\omega}(e^{i\omega} - 1)^2. \label{eq:D2-factor}
\end{align}
Taking the squared modulus and using $|e^{i\theta}| = 1$:
\begin{equation}
\left|[\mathbf{D}_2\,\mathbf{e}_\omega]_s\right|^2
= |e^{i\omega} - 1|^4
= \bigl[(\cos\omega - 1)^2 + \sin^2\omega\bigr]^2
= (2 - 2\cos\omega)^2. \notag
\end{equation}
Since $|e^{i\omega} - 1| = |1 - e^{-i\omega}|$, the two forms in \eqref{eq:D2-gain} are equal.

\paragraph{Corollary 1 (Frequency response of $\mathbf{A}_t$).}
For $\mathbf{A}_t = (1-\lambda_t)\mathbf{I} + \lambda_t\mathbf{D}_2$ with $\lambda_t \in [0, \lambda_{\max}]$, the per-element frequency response at internal positions $s \geq 3$ is
\begin{equation}
H_{\mathbf{A}_t}(\omega) = (1 - \lambda_t) + \lambda_t(1 - e^{-i\omega})^2.
\label{eq:At-freq-response}
\end{equation}
In particular:
\begin{itemize}
\item DC ($\omega = 0$): $|H_{\mathbf{A}_t}(0)|^2 = (1-\lambda_t)^2$, so low-frequency content is attenuated.
\item Nyquist ($\omega = \pi$): $|H_{\mathbf{A}_t}(\pi)|^2 = (1+3\lambda_t)^2$, so high-frequency content is amplified.
\end{itemize}

\paragraph{Proof.}
By linearity, for $s \geq 3$:
\begin{equation}
[\mathbf{A}_t\,\mathbf{e}_\omega]_s
= (1-\lambda_t)\,e^{i(s-1)\omega}
+ \lambda_t\,e^{i(s-3)\omega}(e^{i\omega}-1)^2
= e^{i(s-1)\omega}\!\left[(1-\lambda_t) + \lambda_t\,e^{-2i\omega}(e^{i\omega}-1)^2\right]. \notag
\end{equation}
Since $e^{-2i\omega}(e^{i\omega}-1)^2 = 1 - 2e^{-i\omega} + e^{-2i\omega} = (1-e^{-i\omega})^2$, the bracketed term equals $H_{\mathbf{A}_t}(\omega)$.

At $\omega = 0$: $(1 - e^0)^2 = 0$, giving $H_{\mathbf{A}_t}(0) = 1 - \lambda_t$.

At $\omega = \pi$: $(1 - e^{-i\pi})^2 = (1+1)^2 = 4$, giving $H_{\mathbf{A}_t}(\pi) = 1 + 3\lambda_t$.

Note that $\mathbf{D}_2$ is not a circulant matrix, so $H_{\mathbf{A}_t}(\omega)$ is not an eigenvalue of $\mathbf{A}_t$ but rather the per-element gain for internal rows. Rows $s = 1$ (zero output) and $s = 2$ (first-order difference) have boundary-specific behavior. When $S \gg 1$, the fraction of boundary rows is negligible.

\subsection{Marginal Consistency of the Generalized Reverse Process}
\label{sec:marginal-consistency}

\paragraph{Theorem 1 (Marginal consistency).}
Let the forward marginal be $q(\mathbf{x}_t \mid \mathbf{x}_0) = \mathcal{N}\!\left(\sqrt{\bar{\alpha}_t}\,\mathbf{A}_t\,\mathbf{x}_0,\;(1-\bar{\alpha}_t)\,\mathbf{I}\right)$ with step-dependent transition matrices $\{\mathbf{A}_t\}_{t=0}^{T}$. Consider the reverse update from step $t$ to step $s$ ($s < t$):
\begin{equation}
\mathbf{x}_s
= \sqrt{\bar{\alpha}_s}\,\mathbf{A}_s\,\hat{\mathbf{x}}_0
+ \sqrt{1-\bar{\alpha}_s-\sigma_t^2}\;\frac{\mathbf{x}_t - \sqrt{\bar{\alpha}_t}\,\mathbf{A}_t\,\hat{\mathbf{x}}_0}{\sqrt{1-\bar{\alpha}_t}}
+ \sigma_t\,\boldsymbol{\epsilon}',
\label{eq:reverse-general}
\end{equation}
where $\boldsymbol{\epsilon}' \sim \mathcal{N}(\mathbf{0},\mathbf{I})$ is independent of $\mathbf{x}_t$ and $\sigma_t^2 \leq 1 - \bar{\alpha}_s$. If the denoiser is perfect, i.e., $\hat{\mathbf{x}}_0 = \mathbf{x}_0$, then $\mathbf{x}_s$ has the correct forward marginal:
\begin{equation}
\mathbf{x}_s \sim q(\mathbf{x}_s \mid \mathbf{x}_0) = \mathcal{N}\!\left(\sqrt{\bar{\alpha}_s}\,\mathbf{A}_s\,\mathbf{x}_0,\;(1-\bar{\alpha}_s)\,\mathbf{I}\right).
\end{equation}

\paragraph{Proof.}
Under $\hat{\mathbf{x}}_0 = \mathbf{x}_0$, write $\mathbf{x}_t$ using the forward marginal:
\begin{equation}
\mathbf{x}_t = \sqrt{\bar{\alpha}_t}\,\mathbf{A}_t\,\mathbf{x}_0 + \sqrt{1-\bar{\alpha}_t}\,\boldsymbol{\epsilon},
\qquad \boldsymbol{\epsilon} \sim \mathcal{N}(\mathbf{0},\mathbf{I}).
\label{eq:mc-xt}
\end{equation}
Substituting \eqref{eq:mc-xt} and $\hat{\mathbf{x}}_0 = \mathbf{x}_0$ into \eqref{eq:reverse-general}:
\begin{align}
\mathbf{x}_s
&= \sqrt{\bar{\alpha}_s}\,\mathbf{A}_s\,\mathbf{x}_0
+ \sqrt{1-\bar{\alpha}_s-\sigma_t^2}\;\frac{\sqrt{1-\bar{\alpha}_t}\,\boldsymbol{\epsilon}}{\sqrt{1-\bar{\alpha}_t}}
+ \sigma_t\,\boldsymbol{\epsilon}' \notag\\
&= \sqrt{\bar{\alpha}_s}\,\mathbf{A}_s\,\mathbf{x}_0
+ \sqrt{1-\bar{\alpha}_s-\sigma_t^2}\;\boldsymbol{\epsilon}
+ \sigma_t\,\boldsymbol{\epsilon}'. \label{eq:mc-xs}
\end{align}
The term $\mathbf{A}_t$ cancels entirely: $\mathbf{x}_t - \sqrt{\bar{\alpha}_t}\,\mathbf{A}_t\,\mathbf{x}_0 = \sqrt{1-\bar{\alpha}_t}\,\boldsymbol{\epsilon}$, which is independent of $\mathbf{A}_t$.

Since $\boldsymbol{\epsilon}$ and $\boldsymbol{\epsilon}'$ are independent standard Gaussians, $\mathbf{x}_s$ conditioned on $\mathbf{x}_0$ is Gaussian with
\begin{align}
\mathbb{E}[\mathbf{x}_s \mid \mathbf{x}_0] &= \sqrt{\bar{\alpha}_s}\,\mathbf{A}_s\,\mathbf{x}_0, \notag\\
\mathrm{Cov}[\mathbf{x}_s \mid \mathbf{x}_0] &= \left(1-\bar{\alpha}_s-\sigma_t^2+\sigma_t^2\right)\mathbf{I} = (1-\bar{\alpha}_s)\,\mathbf{I}. \notag
\end{align}
This matches $q(\mathbf{x}_s \mid \mathbf{x}_0)$ exactly. Since $\mathbf{A}_s$ and $\mathbf{A}_t$ are arbitrary (they need not be related), the result holds for any sequence of transition matrices.

The proof shows that marginal consistency holds for \emph{any} choice of step-dependent transition matrices $\{\mathbf{A}_t\}$, not only the specific $\mathbf{A}_t = (1-\lambda_t)\mathbf{I} + \lambda_t\mathbf{D}_2$ used by \textsc{DiffDiff}. The only requirement is that the reverse update \eqref{eq:reverse-general} uses the correct $\mathbf{A}_t$ and $\mathbf{A}_s$ at their respective steps. Setting $\mathbf{A}_t = \mathbf{I}$ for all $t$ recovers the standard DDIM marginal consistency result.

\subsection{Training Objective Derivation}
\label{sec:training-objective}

We derive the training objective used by \textsc{DiffDiff}, starting from the denoising formulation and explaining why $\mathbf{x}_0$-prediction is the natural parameterization for our generalized forward process.

\paragraph{From denoising to reconstruction loss.}
The reverse process in \eqref{eq:ddim-reverse} is fully determined by the denoiser output $\hat{\mathbf{x}}_0 = f_\theta(\mathbf{x}_t, t, \mathbf{x}_{\text{hist}})$. By Theorem 1, perfect denoising ($\hat{\mathbf{x}}_0 = \mathbf{x}_0$) guarantees exact marginal recovery at every reverse step. The natural training objective therefore minimizes the denoising error:
\begin{equation}
\mathcal{L}_{\mathrm{denoise}}
= \mathbb{E}_{t,\,\mathbf{x}_0,\,\boldsymbol{\epsilon}}
\!\left[\left\|\mathbf{x}_0 - f_\theta(\mathbf{x}_t, t, \mathbf{x}_{\text{hist}})\right\|^2\right],
\qquad t \sim \mathrm{Uniform}\{0,\ldots,T{-}1\}.
\label{eq:denoise-obj}
\end{equation}

\paragraph{Why $\mathbf{x}_0$-prediction is task-aligned.}
For the standard forward process ($\mathbf{A}_t = \mathbf{I}$), $\boldsymbol{\epsilon}$-prediction and $\mathbf{x}_0$-prediction are equivalent up to a scalar rescaling: $\|\boldsymbol{\epsilon} - \hat{\boldsymbol{\epsilon}}\|^2 = \frac{\bar{\alpha}_t}{1-\bar{\alpha}_t}\|\mathbf{x}_0 - \hat{\mathbf{x}}_0\|^2$. For our generalized forward process, this equivalence breaks. Given $\mathbf{x}_t = \sqrt{\bar{\alpha}_t}\,\mathbf{A}_t\,\mathbf{x}_0 + \sqrt{1-\bar{\alpha}_t}\,\boldsymbol{\epsilon}$ and $\hat{\boldsymbol{\epsilon}} = (\mathbf{x}_t - \sqrt{\bar{\alpha}_t}\,\mathbf{A}_t\,\hat{\mathbf{x}}_0)/\sqrt{1-\bar{\alpha}_t}$, the noise-matching error becomes
\begin{equation}
\|\boldsymbol{\epsilon} - \hat{\boldsymbol{\epsilon}}\|^2
= \frac{\bar{\alpha}_t}{1-\bar{\alpha}_t}\,\left\|\mathbf{A}_t(\mathbf{x}_0 - \hat{\mathbf{x}}_0)\right\|^2
= \frac{\bar{\alpha}_t}{1-\bar{\alpha}_t}\,(\mathbf{x}_0 - \hat{\mathbf{x}}_0)^\top\mathbf{A}_t^\top\mathbf{A}_t\,(\mathbf{x}_0 - \hat{\mathbf{x}}_0).
\label{eq:eps-to-x0}
\end{equation}
The quadratic form $\mathbf{A}_t^\top\mathbf{A}_t$ introduces a position-coupling that weights prediction errors non-uniformly. Under the interior-symbol approximation (Corollary 1, valid for positions $s \geq 3$), $\mathbf{A}_t$ amplifies high-frequency components (gain $(1+3\lambda_t)^2$ at Nyquist) and attenuates low-frequency ones (gain $(1-\lambda_t)^2$ at DC). The forward operator $\mathbf{A}_t$ already encodes the predictability asymmetry by reshaping the noisy intermediate states themselves, so a loss-side frequency reweighting would compound this inductive bias rather than complement it, while standard evaluation metrics (MSE, CRPS) weight all frequencies uniformly in the value domain. We therefore confine the frequency selectivity to the forward process and adopt the $\mathbf{x}_0$-prediction parameterization, which directly minimizes the value-domain error $\|\mathbf{x}_0 - f_\theta\|^2$ and stays aligned with the downstream evaluation. The noise estimate needed for the reverse step is recovered from the value-domain prediction via \eqref{eq:eps-hat} without requiring matrix inversion.

Under $\mathbf{x}_0$-prediction, the noise-matching error at step $t$ takes the form
\begin{equation}
\|\boldsymbol{\epsilon} - \hat{\boldsymbol{\epsilon}}\|^2
= \frac{\bar{\alpha}_t}{1-\bar{\alpha}_t}\,
\left\|\mathbf{A}_t(\mathbf{x}_0 - f_\theta(\mathbf{x}_t, t, \mathbf{x}_{\text{hist}}))\right\|^2,
\label{eq:weighted-x0-loss}
\end{equation}
where the quadratic form $\|\mathbf{A}_t\boldsymbol{\Delta}\|^2 = \boldsymbol{\Delta}^\top\mathbf{A}_t^\top\mathbf{A}_t\,\boldsymbol{\Delta}$ couples different positions of the prediction error through the transition matrix. The simplified objective in \eqref{eq:denoise-obj} drops both the timestep-dependent scalar $\bar{\alpha}_t/(1-\bar{\alpha}_t)$ and the matrix weighting $\mathbf{A}_t^\top\mathbf{A}_t$, treating all positions and timesteps uniformly. This simplification is analogous to the unweighted $\boldsymbol{\epsilon}$-prediction loss \cite{ho2020denoising} and is justified empirically by improved sample quality.

\paragraph{Instance normalization and position weighting.}
To focus the denoiser on temporal dynamics rather than sample-specific level and scale, the extended target $\mathbf{x}_0 = [\mathbf{x}_{\text{label}};\,\mathbf{x}_{\text{fut}}]$ is instance-normalized along the time dimension~\cite{kim2022revin}:
\begin{equation}
\tilde{\mathbf{x}}_0 = \frac{\mathbf{x}_0 - \mu\,\mathbf{1}}{\nu},
\qquad
\mu = \mathrm{mean}(\mathbf{x}_0),
\quad
\nu = \sqrt{\mathrm{Var}(\mathbf{x}_0) + \epsilon_{\mathrm{norm}}},
\label{eq:instance-norm}
\end{equation}
where $\mu, \nu \in \mathbb{R}$ are per-instance scalars broadcast to all $S$ positions, $\epsilon_{\mathrm{norm}} > 0$ is a small constant ensuring $\nu > 0$ even for near-constant windows (we use $\epsilon_{\mathrm{norm}} = 10^{-6}$), and the forward process operates on $\tilde{\mathbf{x}}_0$. The position-weighted reconstruction loss in \eqref{eq:loss-recon} assigns weight $w_i = w_{\mathrm{fut}} \geq 1$ to future positions and $w_i = 1$ to label positions, directing optimization toward the forecasting segment.

\paragraph{Auxiliary statistics prediction.}
Since the denoiser operates in the normalized space, an auxiliary head predicts $(\hat{\mu}, \hat{\nu})$ from the conditioning feature $\mathbf{c}$. These predictions are trained with MSE losses and used at inference to map the denoised output back to the original scale. The total training objective is
\begin{equation}
\mathcal{L} = \mathcal{L}_{\mathrm{recon}} + \|\hat{\mu} - \mu\|^2 + \|\hat{\nu} - \nu\|^2,
\end{equation}
reproducing \eqref{eq:loss} in the main text.

\subsection{Hybrid Loss Upper Bound}
\label{sec:hybrid-loss}

\paragraph{Proposition 2 (Hybrid loss as upper bound).}
Let $\tilde{\mathbf{x}}_0 = (\mathbf{x}_0 - \mu)/\nu$ be the instance-normalized target where $\mu, \nu \in \mathbb{R}$ are the per-instance mean and standard deviation, and let $(\hat{\mu}, \hat{\nu})$ be the predicted statistics. The unnormalized reconstruction error satisfies
\begin{equation}
\left\|\hat{\nu}\,f_\theta(\mathbf{x}_t,t,\mathbf{x}_{\text{hist}}) + \hat{\mu}\,\mathbf{1} - \mathbf{x}_0\right\|^2
\;\leq\;
3\left(\hat{\nu}^2\,\mathcal{R} + (\hat{\nu} - \nu)^2\,\|\tilde{\mathbf{x}}_0\|^2 + S\,(\hat{\mu} - \mu)^2\right),
\label{eq:hybrid-bound}
\end{equation}
where $\mathcal{R} = \|f_\theta(\mathbf{x}_t,t,\mathbf{x}_{\text{hist}}) - \tilde{\mathbf{x}}_0\|^2$ is the normalized reconstruction error.

\paragraph{Proof.}
Write the unnormalized prediction as $\hat{\mathbf{x}}_0^{\text{unnorm}} = \hat{\nu}\,f_\theta + \hat{\mu}\,\mathbf{1}$ and the true target as $\mathbf{x}_0 = \nu\,\tilde{\mathbf{x}}_0 + \mu\,\mathbf{1}$. Decompose the error:
\begin{align}
\hat{\mathbf{x}}_0^{\text{unnorm}} - \mathbf{x}_0
&= \hat{\nu}\,f_\theta + \hat{\mu}\,\mathbf{1} - \nu\,\tilde{\mathbf{x}}_0 - \mu\,\mathbf{1} \notag\\
&= \hat{\nu}\,(f_\theta - \tilde{\mathbf{x}}_0)
+ (\hat{\nu} - \nu)\,\tilde{\mathbf{x}}_0
+ (\hat{\mu} - \mu)\,\mathbf{1}. \label{eq:three-term}
\end{align}
Apply $\|\mathbf{a} + \mathbf{b} + \mathbf{c}\|^2 \leq 3(\|\mathbf{a}\|^2 + \|\mathbf{b}\|^2 + \|\mathbf{c}\|^2)$ (from convexity of $\|\cdot\|^2$ and Jensen's inequality) to bound each term:
\begin{align}
\|\hat{\nu}\,(f_\theta - \tilde{\mathbf{x}}_0)\|^2
&= \hat{\nu}^2\,\|f_\theta - \tilde{\mathbf{x}}_0\|^2
= \hat{\nu}^2\,\mathcal{R}, \notag\\
\|(\hat{\nu} - \nu)\,\tilde{\mathbf{x}}_0\|^2
&= (\hat{\nu} - \nu)^2\,\|\tilde{\mathbf{x}}_0\|^2, \notag\\
\|(\hat{\mu} - \mu)\,\mathbf{1}\|^2
&= S\,(\hat{\mu} - \mu)^2. \notag
\end{align}
Combining gives \eqref{eq:hybrid-bound}.

\paragraph{Connection to the weighted training loss.}
The bound in \eqref{eq:hybrid-bound} uses the unweighted reconstruction error $\mathcal{R} = \|f_\theta - \tilde{\mathbf{x}}_0\|^2$. Since the actual training loss employs position weights $w_i \geq 1$ (with $w_i = w_{\mathrm{fut}} \geq 1$ for future positions and $w_i = 1$ for label positions), the per-instance weighted loss $\mathcal{L}_{\mathrm{recon}}^{\mathrm{inst}} = \sum_i w_i(f_{\theta,i} - \tilde{x}_{0,i})^2$ satisfies $\mathcal{R} \leq \mathcal{L}_{\mathrm{recon}}^{\mathrm{inst}}$, so minimizing $\mathcal{L}_{\mathrm{recon}}$ also drives $\mathcal{R} \to 0$. Combined with the auxiliary losses driving $\hat{\mu} \to \mu$ and $\hat{\nu} \to \nu$, the three-term objective jointly controls the unnormalized prediction error.

\paragraph{Boundedness of $\hat{\nu}$.}
The factor $\hat{\nu}^2$ in the first term of \eqref{eq:hybrid-bound} requires that $\hat{\nu}$ be bounded for the bound to be operationally meaningful. In our implementation, $\hat{\nu}$ is produced by a linear projection without activation, so it is not constrained to be positive. We rely on the auxiliary MSE loss $\|\hat{\nu} - \nu\|^2$ to drive $\hat{\nu}$ toward the true $\nu > 0$ (which is guaranteed positive by the $\epsilon_{\mathrm{norm}}$ stabilization in \eqref{eq:instance-norm}), ensuring the bound remains tight during training.

\subsection{Frequency-Dependent Signal-to-Noise Ratio}
\label{sec:freq-snr}

This section analyzes how the forward process design of \textsc{DiffDiff} creates a frequency-dependent signal-to-noise ratio (SNR) profile that differs qualitatively from the isotropic corruption of standard diffusion.

\paragraph{Per-frequency SNR definition.}
Consider the forward marginal $q(\mathbf{x}_t \mid \mathbf{x}_0) = \mathcal{N}(\sqrt{\bar{\alpha}_t}\,\mathbf{A}_t\,\mathbf{x}_0,\,(1-\bar{\alpha}_t)\mathbf{I})$. The signal component at step $t$ is $\sqrt{\bar{\alpha}_t}\,\mathbf{A}_t\,\mathbf{x}_0$ and the noise variance is $(1-\bar{\alpha}_t)$ per dimension. Since $\mathbf{D}_2$ is not a circulant matrix, discrete Fourier modes are not exact eigenvectors of $\mathbf{A}_t$. However, for the interior positions ($s \geq 3$), which constitute a fraction $(S{-}2)/S$ of the sequence, the operator acts as a shift-invariant filter with frequency response $H_{\mathbf{A}_t}(\omega)$ from Corollary 1. For $S \gg 1$, we therefore analyze the \emph{effective} per-frequency signal power in the interior of $\mathbf{A}_t\mathbf{x}_0$ as being modulated by $|H_{\mathbf{A}_t}(\omega)|^2$ relative to the original signal power $|X_0(\omega)|^2$. This motivates the effective per-frequency SNR:
\begin{equation}
\mathrm{SNR}_{\mathrm{eff}}(\omega, t)
= \frac{\bar{\alpha}_t\,|H_{\mathbf{A}_t}(\omega)|^2\,|X_0(\omega)|^2}{1-\bar{\alpha}_t}.
\label{eq:per-freq-snr}
\end{equation}
This definition captures the dominant behavior for long sequences; boundary corrections affect only the first two positions and become negligible as $S$ grows.

\paragraph{Standard diffusion ($\mathbf{A}_t = \mathbf{I}$).}
When $\mathbf{A}_t = \mathbf{I}$ for all $t$, $|H_{\mathbf{I}}(\omega)|^2 = 1$ at all frequencies (with no boundary issue since $\mathbf{I}$ is exactly Fourier-diagonal), so
\begin{equation}
\mathrm{SNR}_{\mathrm{eff,std}}(\omega, t) = \frac{\bar{\alpha}_t\,|X_0(\omega)|^2}{1-\bar{\alpha}_t}.
\end{equation}
All frequency components share the same SNR decay profile $\bar{\alpha}_t/(1-\bar{\alpha}_t)$, scaled only by the signal power $|X_0(\omega)|^2$. The isotropic forward process treats low-frequency and high-frequency information identically.

\paragraph{\textsc{DiffDiff} ($\mathbf{A}_t = (1{-}\lambda_t)\mathbf{I} + \lambda_t\mathbf{D}_2$).}
Using the frequency response from Corollary 1:
\begin{equation}
\mathrm{SNR}_{\mathrm{eff,DD}}(\omega, t) = \frac{\bar{\alpha}_t\,\left|(1-\lambda_t) + \lambda_t(1-e^{-i\omega})^2\right|^2\,|X_0(\omega)|^2}{1-\bar{\alpha}_t}.
\end{equation}

\paragraph{Proposition 3 (Frequency-selective SNR modulation, interior approximation).}
Under the interior symbol approximation (valid for positions $s \geq 3$; see Proposition 1), the ratio of \textsc{DiffDiff} to standard effective per-frequency SNR is
\begin{equation}
\frac{\mathrm{SNR}_{\mathrm{eff,DD}}(\omega, t)}{\mathrm{SNR}_{\mathrm{eff,std}}(\omega, t)}
= |H_{\mathbf{A}_t}(\omega)|^2
= \left|(1-\lambda_t) + \lambda_t(1-e^{-i\omega})^2\right|^2.
\label{eq:snr-ratio}
\end{equation}
This ratio satisfies:
\begin{enumerate}
\item At DC ($\omega = 0$): $\mathrm{SNR}_{\mathrm{eff,DD}} / \mathrm{SNR}_{\mathrm{eff,std}} = (1-\lambda_t)^2 < 1$ for $\lambda_t \in (0,1)$.
\item At Nyquist ($\omega = \pi$): $\mathrm{SNR}_{\mathrm{eff,DD}} / \mathrm{SNR}_{\mathrm{eff,std}} = (1+3\lambda_t)^2 > 1$ for $\lambda_t > 0$.
\item For $\lambda_t \in (0,1)$, there exists a unique crossover frequency $\omega^* \in (0, \pi)$ where $|H_{\mathbf{A}_t}(\omega^*)|^2 = 1$.
\end{enumerate}

\paragraph{Proof.}
Items (1) and (2) follow directly from Corollary 1. For item (1), $(1-\lambda_t)^2 < 1$ holds whenever $\lambda_t \in (0,1)$ (equivalently $0 < 1-\lambda_t < 1$). For item (3), we have $|H_{\mathbf{A}_t}(0)|^2 = (1-\lambda_t)^2 < 1$ and $|H_{\mathbf{A}_t}(\pi)|^2 = (1+3\lambda_t)^2 > 1$ for $\lambda_t \in (0,1)$. Since $|H_{\mathbf{A}_t}(\omega)|^2$ is a continuous function of $\omega$ (as a composition of continuous functions), the intermediate value theorem guarantees the existence of $\omega^* \in (0,\pi)$ with $|H_{\mathbf{A}_t}(\omega^*)|^2 = 1$. To show uniqueness, substitute $c = 1 - \cos\omega \in [0, 2]$ and expand $|H_{\mathbf{A}_t}(\omega)|^2 = 1$ into the quadratic $4c^2 - 4(1{-}\lambda_t)c - (2{-}\lambda_t) = 0$. The two roots are $c = \bigl[(1{-}\lambda_t) \pm \sqrt{(1{-}\lambda_t)^2 + (2{-}\lambda_t)}\bigr]/2$; the ``$-$'' root is negative (since the discriminant exceeds $1{-}\lambda_t$ for $\lambda_t \in (0,1)$), so exactly one root lies in $(0,2)$, giving a unique $\omega^* = \arccos(1-c)$. In practice, $\lambda_t \leq \lambda_{\max} = 0.5 < 1$, so the condition is always satisfied.

\paragraph{Forecasting interpretation.}
The SNR ratio in \eqref{eq:snr-ratio} reveals the inductive bias introduced by the differencing forward process:
\begin{itemize}
\item \emph{Low-frequency suppression} ($\omega$ near $0$). Level and trend components see their SNR reduced by a factor $(1-\lambda_t)^2$ relative to standard diffusion. As $\lambda_t$ increases with the diffusion step, these components are destroyed faster, reflecting the assumption that they can be anchored by the observed history.
\item \emph{High-frequency preservation} ($\omega$ near $\pi$). Temporal dynamics, fluctuations, and local change patterns see their SNR amplified by $(1+3\lambda_t)^2$. These components persist longer in the noisy intermediate states, giving the denoiser more signal about short-term dynamics at later reverse steps.
\item \emph{Progressive transition}. Since $\lambda_t$ increases monotonically from $0$ to $\lambda_{\max}$, the frequency selectivity grows with the diffusion step. Early steps ($\lambda_t \approx 0$) remain nearly isotropic, while later steps emphasize the frequency separation.
\end{itemize}
This frequency-dependent SNR profile aligns with the forecasting-specific division of labor: the history already anchors low-frequency structure, so the generative process can focus its capacity on the harder-to-predict high-frequency dynamics.

The above analysis uses the interior symbol $H_{\mathbf{A}_t}(\omega)$, which is exact only for positions $s \geq 3$ of the finite operator $\mathbf{A}_t$. For a length-$S$ sequence, boundary effects at $s \in \{1, 2\}$ introduce deviations from the idealized frequency response. In practice, with $S \geq 48$ (the shortest target sequence in our experiments), boundary positions constitute at most $\approx 4.2\%$ of the sequence. Furthermore, the noise covariance remains exactly $(1-\bar{\alpha}_t)\mathbf{I}$ regardless of $\mathbf{A}_t$ (which modulates only the signal mean, not the noise), so the flat noise spectrum assumed in the SNR definition is exact. The interior symbol therefore provides an accurate characterization of the dominant SNR behavior for long sequences.

\subsection{Terminal Distribution Convergence}
\label{sec:terminal-convergence}

\paragraph{Proposition 4 (Terminal convergence).}
For any $\mathbf{x}_0$ with $\|\mathbf{x}_0\| < \infty$ and any sequence of matrices $\{\mathbf{A}_t\}$ with uniformly bounded operator norm $\|\mathbf{A}_t\|_{\mathrm{op}} \leq C$ for all $t$,
\begin{equation}
D_{\mathrm{KL}}\!\left(q(\mathbf{x}_T \mid \mathbf{x}_0)\;\|\;\mathcal{N}(\mathbf{0},\mathbf{I})\right)
\;\to\; 0
\qquad \text{as } \bar{\alpha}_T \to 0.
\end{equation}

\paragraph{Proof.}
For $q(\mathbf{x}_T \mid \mathbf{x}_0) = \mathcal{N}(\boldsymbol{\mu}_T, \boldsymbol{\Sigma}_T)$ with $\boldsymbol{\mu}_T = \sqrt{\bar{\alpha}_T}\,\mathbf{A}_T\mathbf{x}_0$ and $\boldsymbol{\Sigma}_T = (1-\bar{\alpha}_T)\mathbf{I}$, the KL divergence to the standard Gaussian is
\begin{equation}
D_{\mathrm{KL}} = \frac{1}{2}\!\left[\mathrm{tr}(\boldsymbol{\Sigma}_T) - S + \|\boldsymbol{\mu}_T\|^2 - \ln\det(\boldsymbol{\Sigma}_T)\right],
\label{eq:kl-terminal}
\end{equation}
where $S$ is the dimensionality. Substituting:
\begin{align}
\mathrm{tr}(\boldsymbol{\Sigma}_T) &= S(1-\bar{\alpha}_T), \notag\\
\ln\det(\boldsymbol{\Sigma}_T) &= S\ln(1-\bar{\alpha}_T), \notag\\
\|\boldsymbol{\mu}_T\|^2 &= \bar{\alpha}_T\,\|\mathbf{A}_T\mathbf{x}_0\|^2 \leq \bar{\alpha}_T\,C^2\|\mathbf{x}_0\|^2. \notag
\end{align}
Therefore
\begin{equation}
D_{\mathrm{KL}}
= \frac{1}{2}\!\left[-S\bar{\alpha}_T + \bar{\alpha}_T\|\mathbf{A}_T\mathbf{x}_0\|^2 - S\ln(1-\bar{\alpha}_T)\right].
\label{eq:kl-expanded}
\end{equation}
As $\bar{\alpha}_T \to 0$, using $\ln(1-\bar{\alpha}_T) = -\bar{\alpha}_T + O(\bar{\alpha}_T^2)$:
\begin{equation}
D_{\mathrm{KL}}
= \frac{1}{2}\!\left[-S\bar{\alpha}_T + \bar{\alpha}_T\|\mathbf{A}_T\mathbf{x}_0\|^2 + S\bar{\alpha}_T + O(\bar{\alpha}_T^2)\right]
= \frac{\bar{\alpha}_T}{2}\|\mathbf{A}_T\mathbf{x}_0\|^2 + O(\bar{\alpha}_T^2).
\end{equation}
Since $\|\mathbf{A}_T\mathbf{x}_0\|^2 \leq C^2\|\mathbf{x}_0\|^2 < \infty$, the KL divergence vanishes as $\bar{\alpha}_T \to 0$.
For the \textsc{DiffDiff} parameterization with $\mathbf{A}_t = (1-\lambda_t)\mathbf{I} + \lambda_t\mathbf{D}_2$, the operator norm satisfies $\|\mathbf{A}_t\|_{\mathrm{op}} \leq (1-\lambda_t) + \lambda_t\|\mathbf{D}_2\|_{\mathrm{op}}$. Since $\mathbf{D}_2$ is a fixed finite-dimensional matrix, $\|\mathbf{D}_2\|_{\mathrm{op}}$ is bounded, confirming the uniform bound assumption.

\subsection{Extended Target and Cross-Boundary Information}
\label{sec:extended-target-theory}

The extended target $\mathbf{x}_0 = [\mathbf{x}_{\text{label}};\,\mathbf{x}_{\text{fut}}] \in \mathbb{R}^{S}$ with $S = L_l + H$ includes a label window of length $L_l \geq 2$ from the end of the observed history. This section formalizes why the label overlap is beneficial under the differencing forward process.

\paragraph{Proposition 5 (Cross-boundary information in extended vs.\ future-only targets).}
Let $\mathbf{x}_{\text{fut}} = [x_1^f, \ldots, x_H^f]^\top \in \mathbb{R}^H$ be the future window and $\mathbf{x}_{\text{ext}} = [x_{-L_l+1}, \ldots, x_0, x_1^f, \ldots, x_H^f]^\top \in \mathbb{R}^S$ be the extended target where $x_0$ is the last observed value. Under $\mathbf{D}_2$ as defined in \eqref{eq:D2}:
\begin{enumerate}
\item The future-only target satisfies $[\mathbf{D}_2^{(H)} \mathbf{x}_{\text{fut}}]_1 = 0$ and $[\mathbf{D}_2^{(H)} \mathbf{x}_{\text{fut}}]_2 = x_2^f - x_1^f$, encoding only differences within the future window.
\item The extended target satisfies $[\mathbf{D}_2^{(S)} \mathbf{x}_{\text{ext}}]_{L_l+1} = x_1^f - 2x_0 + x_{-1}$, encoding the second-order difference across the history--future boundary.
\end{enumerate}
Consequently, the differencing forward process applied to the extended target preserves cross-boundary curvature information at positions near the junction, which is lost when the target is restricted to the future window alone.

\paragraph{Proof.}
Item (1) follows directly from the definition of $\mathbf{D}_2^{(H)}$ operating on $\mathbb{R}^H$: position 1 outputs zero (removes absolute level), and position 2 outputs $x_2^f - x_1^f$ (first-order difference within the future). The first future value $x_1^f$ and its relationship to the last observed value $x_0$ do not appear anywhere in $\mathbf{D}_2^{(H)} \mathbf{x}_{\text{fut}}$.

For item (2), in the extended target the position corresponding to $x_1^f$ is at index $L_l + 1$ (1-based). Since $L_l + 1 \geq 3$ (given $L_l \geq 2$), the full second-order difference applies:
\begin{equation}
[\mathbf{D}_2^{(S)} \mathbf{x}_{\text{ext}}]_{L_l+1} = x_1^f - 2x_0 + x_{-1},
\end{equation}
which encodes the local curvature at the history--future boundary. Similarly, position $L_l$ encodes $x_0 - 2x_{-1} + x_{-2}$, providing the curvature just before the boundary.

The proposition shows that the label overlap serves a specific structural role beyond general ``boundary anchoring'': it ensures that the differencing operator can compute cross-boundary differences that link the future trajectory to the observed history. Under the forward process $\mathbf{x}_t = \sqrt{\bar{\alpha}_t}\,\mathbf{A}_t\,\mathbf{x}_0 + \sqrt{1-\bar{\alpha}_t}\,\boldsymbol{\epsilon}$, these cross-boundary differences are preserved in the signal component at intermediate diffusion steps (with strength modulated by $\lambda_t$). This provides the denoiser with explicit continuity cues that would be absent if only the future window were used as the diffusion target.

\section{Dataset Statistics}
\label{sec:appendix-datasets}

\begin{table}[h]
\centering
\caption{Dataset statistics.}
\label{tab:datasets}
\small
\begin{tabular}{lrrrrr}
\toprule
Dataset & Frequency & Train & Val & Test \\
\midrule
\texttt{Electricity} & 1 hour & 18,317 & 2,633 & 5,261 \\
\texttt{ETTm2}       & 15 min & 34,465 & 11,521 & 11,521 \\
\texttt{Exchange}    & 1 day  & 5,120  & 665    & 1,422 \\
\texttt{Traffic}     & 1 hour & 12,185 & 1,757  & 3,509 \\
\texttt{Weather}     & 10 min & 36,792 & 5,271  & 10,540 \\
\texttt{Solar}       & 10 min & 36,601 & 5,161  & 10,417 \\
\texttt{Wind}        & 15 min & 38,444 & 4,990  & 10,523 \\
\bottomrule
\end{tabular}
\end{table}

\section{Evaluation Metrics}
\label{sec:appendix-metrics}

Let $y_{i,t}$ denote the ground-truth value and $\hat{y}_{i,t}$ the point prediction for the $i$-th forecast window at time step $t$, with $i = 1, \ldots, N$ and $t = 1, \ldots, H$. The point-prediction metrics are mean squared error (MSE) and mean absolute error (MAE):
\begin{equation}
\mathrm{MSE} = \frac{1}{N H} \sum_{i=1}^{N} \sum_{t=1}^{H} (y_{i,t} - \hat{y}_{i,t})^2,
\qquad
\mathrm{MAE} = \frac{1}{N H} \sum_{i=1}^{N} \sum_{t=1}^{H} \lvert y_{i,t} - \hat{y}_{i,t} \rvert.
\end{equation}

Probabilistic forecast quality is measured by the continuous ranked probability score (CRPS), computed via a multi-quantile approximation with $|Q|=9$ uniformly spaced quantile levels $Q = \{0.1, 0.2, \ldots, 0.9\}$. For each window we draw $K = 100$ samples from the predictive distribution, and let $\hat{y}_{i,t}^{(q)}$ denote the empirical $q$-quantile at position $(i, t)$. Then
\begin{equation}
\mathrm{CRPS} = \frac{2}{N H |Q|} \sum_{i=1}^{N} \sum_{t=1}^{H} \sum_{q \in Q} \rho_q\!\left( y_{i,t} - \hat{y}_{i,t}^{(q)} \right),
\quad
\rho_q(u) = \max\{q u,\, (q-1) u\}.
\end{equation}
Lower values of all three metrics indicate better forecasts.

\section{{Comparison with Deterministic Forecasters}}
\label{sec:appendix-deterministic-baselines}

{Table~\ref{tab:main} compares probabilistic diffusion forecasters under matched distributional objectives. To further evaluate \textsc{DiffDiff}'s point-forecasting performance in a broader forecasting context, we additionally compare it with recent state-of-the-art deterministic forecasters, TQNet, Times2D, and PathFormer, using the same seven datasets, chronological splits, four horizons, and $L=96$ inputs, as shown in Table~\ref{tab:deterministic-baselines}. These deterministic methods directly optimize point-estimation objectives and are therefore typically more competitive in point forecasting. \textsc{DiffDiff} obtains the lowest mean MSE on five datasets and the lowest overall MSE and MAE while retaining a predictive distribution. The gain is therefore not confined to rankings among diffusion models, although the objectives remain different and the deterministic methods have no matched CRPS comparison.}

\begin{table}[t]
\centering
\caption{{Point-forecasting comparison with deterministic forecasters using four-horizon mean MSE and MAE under the shared $L=96$ protocol. Lower is better.}}
\label{tab:deterministic-baselines}
\footnotesize
\setlength{\tabcolsep}{2.8pt}
{
\resizebox{\textwidth}{!}{
\begin{tabular}{lcccccccc}
\toprule
Model & Electricity & ETTm2 & Exchange & Traffic & Weather & Solar & Wind & Overall \\
\midrule
TQNet & .376/.436 & .120/.256 & .552/.493 & .151/.227 & .189/.307 & .881/.603 & 1.107/.762 & .482/.440 \\
Times2D & .362/.428 & .123/.259 & .496/.492 & .171/.258 & .190/.308 & .936/.623 & 1.098/.762 & .482/.447 \\
PathFormer & .348/.417 & .119/.253 & .615/.522 & .162/.238 & .190/.309 & 1.028/.619 & 1.074/.744 & .505/.443 \\
\textsc{DiffDiff} & \textbf{.338/.407} & .120/.258 & \textbf{.356/.457} & \textbf{.156/.237} & .190/.315 & \textbf{.455/.433} & \textbf{.823/.687} & \textbf{.348/.399} \\
\bottomrule
\end{tabular}}}
\end{table}

\section{{Rapid Local Changes as an Indicator of Forecasting Difficulty}}
\label{sec:appendix-forecasting-difficulty}

{To examine whether the rapid local changes emphasized by the differencing operator coincide with future content that is difficult to infer from history, we relate local curvature to predictor error and matched-history dispersion, as summarized in Table~\ref{tab:forecasting-difficulty}. For each forecast position $t$, we define local curvature as}
\begin{equation}
{c_t = \left|y_t - 2y_{t-1} + y_{t-2}\right|.}
\label{eq:local-curvature}
\end{equation}
{Here, $y_t$ is the ground-truth value at position $t$, and $c_t$ measures the absolute change between two consecutive slopes. At the first forecast position, $y_{t-1}$ and $y_{t-2}$ are the final two history observations, matching the cross-boundary construction of \textsc{DiffDiff}. We quantify forecasting difficulty in two complementary ways across the seven datasets and four horizons. Predictor error is the position-wise absolute error $|\hat{y}_t-y_t|$ of TQNet or Times2D, averaged over five seeds. Matched-history dispersion is the standard deviation of futures following similar training histories in a training-only PCA space. The $q50$ and $q75$ variants use distance thresholds at the median and $75$th percentile of training-history distances, respectively, so $q50$ forms the tighter neighborhood. All normalization, PCA fitting, and neighbor selection use training data only.}

{Within each dataset-horizon setting, Spearman's $\rho$ measures the monotonic association between $c_t$ and a difficulty measure. The 95\% confidence interval summarizes uncertainty in the mean correlation across settings, while Positive settings counts the settings with $\rho>0$. Q4/Q1 is the mean difficulty in the highest-curvature quartile divided by that in the lowest-curvature quartile. The matched-history analysis yields valid neighborhoods in 24 settings.}

\begin{table}[t]
\centering
\caption{{Association between local curvature and forecasting difficulty. Predictor statistics report mean $\pm$ sample standard deviation over five seeds.}}
\label{tab:forecasting-difficulty}
\scriptsize
\setlength{\tabcolsep}{2.5pt}
{
\begin{tabular}{lcccc}
\toprule
Difficulty measure & Mean Spearman $\rho$ & 95\% CI & Positive settings & Q4/Q1 \\
\midrule
TQNet absolute error & $.1417\pm.0035$ & [.0851, .2077] & 26/28 & $1.6427\pm.0455$ \\
Times2D absolute error & $.1282\pm.0038$ & [.0749, .1919] & 26/28 & $1.5841\pm.0378$ \\
Matched-history dispersion ($q50$) & .1304 & [.0393, .2310] & 14/24 & 1.8845 \\
Matched-history dispersion ($q75$) & .1331 & [.0504, .2350] & 22/24 & 1.5355 \\
\bottomrule
\end{tabular}}
\end{table}

{Table~\ref{tab:forecasting-difficulty} shows that both deterministic predictors produce positive correlations in 26 of 28 settings, with roughly $1.6\times$ larger errors in the highest-curvature quartile than in the lowest. The model-free matched-history comparison preserves the positive mean association at both neighborhood widths, showing that rapid local changes also coincide with greater variation among futures following similar histories. The agreement between predictor errors and matched-history dispersion identifies local curvature as a consistent indicator of history-conditioned forecasting difficulty. The differencing operator therefore emphasizes locally difficult dynamics in accordance with the predictability asymmetry underlying \textsc{DiffDiff}.}

\section{Conditioning Mechanism Variants}
\label{sec:appendix-variants}
To verify the necessity of each design choice, we compare \textsc{DiffDiff} against four ablation variants, each replacing one component of this pathway.

\paragraph{Variant descriptions.}
All variants share the same forward operator $\mathbf{A}_t = (1{-}\lambda_t)\mathbf{I} + \lambda_t \mathbf{D}_2$, the same value-domain main encoding $\mathbf{h}_{\text{val}} = \text{MLP}_{\text{val}}(\mathbf{x}_{\text{hist}})$, and the same denoiser backbone. They differ only in how the differential signal is incorporated into the conditioning feature.

\begin{itemize}[leftmargin=1.5em,itemsep=2pt]
\item \textbf{CrossAttn.}
The sigmoid gate is replaced by a single-head cross-attention layer~\cite{vaswani2017attention} that selects between two views, the value-domain encoding $\bar{\mathbf{h}}_{\text{val}}$ and the differential encoding $\bar{\mathbf{h}}_{\text{diff}}$ (both channel-averaged). The query is projected from $[\mathbf{t}_{\text{emb}};\, \bar{\mathbf{h}}_{\text{val}}]$, and the keys and values are the two view tokens. The softmax produces a convex combination of the two views, in contrast to the unconstrained per-dimension sigmoid in \textsc{DiffDiff}.

\item \textbf{OpenGate.}
The same architecture as \textsc{DiffDiff} except the gate bias is set to a large positive value ($\sigma{\approx}1$), so the differential branch is always fully applied without timestep-dependent modulation. This isolates whether the gate's selective-suppression ability is necessary.

\item \textbf{DualOrder.}
A second-order differential branch encoding $[\Delta^2 \mathbf{x}_{\text{hist}}]_\tau = \mathbf{x}_{\text{hist},\tau} - 2\mathbf{x}_{\text{hist},\tau-1} + \mathbf{x}_{\text{hist},\tau-2}$ (with $\tau$ indexing positions within the history window) is added alongside the first-order branch, each with its own independent sigmoid gate: $\mathbf{h} = \mathbf{h}_{\text{val}} + \mathbf{g}_1 \odot \mathbf{h}_{\Delta^1} + \mathbf{g}_2 \odot \mathbf{h}_{\Delta^2}$. This tests whether the conditioning pathway benefits from explicit second-order differential information beyond the first-order branch.

\item \textbf{Concat.}
The sigmoid gate is removed and the two branches are fused by concatenation followed by a small MLP: $\mathbf{c} = \mathrm{MLP}_{\mathrm{fuse}}([\mathbf{h}_{\text{val}};\, \mathbf{h}_{\text{diff}}])$. This tests whether the gain attributed to the stage-adaptive gate could be explained by the additional model capacity introduced when both branches are present, rather than by the selective-admission mechanism itself.
\end{itemize}

Table~\ref{tab:variants} compares all five variants on the seven main benchmarks across four prediction horizons (five-seed average). \textsc{DiffDiff} obtains 54 of 84 wins, supporting the chosen design.

Replacing the sigmoid gate with cross-attention (\textbf{CrossAttn}) introduces a softmax-induced zero-sum trade-off, where the convex combination must shift mass away from the value view to admit the differential signal. CrossAttn wins concentrate on Exchange at short horizons and on Solar and Wind at long horizons, but the variant remains behind \textsc{DiffDiff} on the strongly periodic and stationary benchmarks where the per-dimension sigmoid preserves both views simultaneously.

Forcing the gate fully open (\textbf{OpenGate}) removes the model's ability to suppress the differential branch when its signal is uninformative, exposing the gate's selective-suppression role as the source of cross-regime robustness.

Adding a second-order branch (\textbf{DualOrder}) doubles the gating optimization load without consistent gain, with the variant retaining only one win at $H{=}720$, consistent with extra capacity introducing optimization difficulty rather than additional usable structure.

Replacing the gate with concatenation-and-MLP fusion (\textbf{Concat}) introduces strictly more parameters than \textsc{DiffDiff} yet collapses to two wins, mirroring OpenGate's failure mode and ruling out larger fusion capacity as the source of \textsc{DiffDiff}'s gain. The benefit thus stems specifically from the gate's ability to suppress the differential branch when its signal is uninformative.

A single first-order differential branch fused with the value-domain encoding through a learned sigmoid gate suffices to deliver the predictability-aligned inductive bias while preserving robustness on data where differential information is uninformative.

\begin{table}[t]
\centering
\caption{Conditioning ablation. Within each (dataset, metric) column the best value across variants is in \textbf{bold} and the second-best is \underline{underlined}. The rightmost column reports per-row wins across the 21 (dataset, metric) settings.}
\label{tab:variants}
\footnotesize
\setlength{\tabcolsep}{2.0pt}
\resizebox{\textwidth}{!}{
\begin{tabular}{c|c| ccc| ccc| ccc| ccc| ccc| ccc| ccc| |c}
\toprule
\multirow{2}{*}{Variant} & \multirow{2}{*}{$H$}
 & \multicolumn{3}{c|}{Electricity}
 & \multicolumn{3}{c|}{ETTm2}
 & \multicolumn{3}{c|}{Exchange}
 & \multicolumn{3}{c|}{Traffic}
 & \multicolumn{3}{c|}{Weather}
 & \multicolumn{3}{c|}{Solar}
 & \multicolumn{3}{c||}{Wind}
 & \multirow{2}{*}{Wins} \\
\cmidrule(lr){3-5} \cmidrule(lr){6-8} \cmidrule(lr){9-11} \cmidrule(lr){12-14} \cmidrule(lr){15-17} \cmidrule(lr){18-20} \cmidrule(lr){21-23}
 & & MSE & MAE & CRPS & MSE & MAE & CRPS & MSE & MAE & CRPS & MSE & MAE & CRPS & MSE & MAE & CRPS & MSE & MAE & CRPS & MSE & MAE & CRPS & \\
\midrule
\multirow{4}{*}{\rotatebox{90}{\textsc{DiffDiff}}}
 & 96 & \bf .269 & \bf .362 & \bf .150 & .068 & .190 & \underline{.079} & .101 & .246 & .103 & \bf .156 & \bf .234 & \bf .108 & \bf .093 & \bf .222 & \bf .093 & \bf .463 & \bf .435 & \bf .191 & \bf .599 & \bf .547 & \bf .226 & 15 \\
 & 192 & \bf .299 & \bf .381 & \bf .158 & .102 & .239 & .098 & \underline{.198} & \underline{.358} & \underline{.147} & \bf .151 & \bf .231 & \bf .106 & \bf .135 & \bf .271 & \bf .112 & .483 & .448 & .191 & \bf .815 & \bf .675 & \underline{.278} & 11 \\
 & 336 & \bf .349 & \bf .413 & \bf .171 & \bf .128 & \bf .273 & \bf .110 & .388 & .495 & .207 & \bf .148 & \bf .232 & \bf .106 & \bf .202 & \bf .332 & \bf .137 & \underline{.389} & \underline{.395} & \underline{.163} & \bf .912 & \bf .742 & \bf .300 & 15 \\
 & 720 & \bf .435 & \bf .474 & \bf .196 & \underline{.182} & \underline{.330} & \underline{.133} & \bf .735 & \bf .728 & \bf .327 & \bf .168 & \bf .252 & \bf .114 & \bf .328 & \bf .434 & \bf .178 & \underline{.485} & .455 & \underline{.189} & \bf .966 & \underline{.785} & \underline{.317} & 13 \\
\cmidrule(lr){1-24}
\multirow{4}{*}{\rotatebox{90}{CrossAttn}}
 & 96 & \underline{.284} & \underline{.372} & \underline{.154} & .068 & .191 & \underline{.079} & \bf .096 & \bf .238 & \bf .098 & .158 & .236 & .110 & \underline{.094} & \underline{.224} & .095 & \underline{.467} & \underline{.436} & \underline{.192} & .611 & \underline{.548} & \bf .226 & 4 \\
 & 192 & .313 & .390 & \underline{.160} & .102 & .239 & \underline{.097} & \bf .190 & \bf .346 & \bf .141 & .154 & \underline{.233} & \underline{.108} & .139 & .277 & \underline{.114} & .472 & .446 & .189 & .827 & \bf .675 & \bf .273 & 5 \\
 & 336 & .362 & .425 & .174 & .133 & .278 & .113 & \underline{.369} & \underline{.488} & \underline{.203} & .153 & .238 & .112 & .215 & .346 & .143 & \bf .384 & \bf .394 & \bf .162 & .927 & .750 & .306 & 3 \\
 & 720 & \underline{.443} & \underline{.483} & \underline{.199} & .189 & .335 & .136 & .808 & .766 & .347 & .171 & .257 & .119 & \underline{.334} & \underline{.436} & \underline{.179} & \bf .476 & \bf .453 & \bf .187 & \underline{.974} & \bf .782 & \bf .308 & 5 \\
\cmidrule(lr){1-24}
\multirow{4}{*}{\rotatebox{90}{OpenGate}}
 & 96 & .287 & .373 & .156 & \underline{.067} & \underline{.189} & \bf .078 & \underline{.100} & \underline{.245} & \underline{.102} & .159 & .236 & \underline{.109} & \underline{.094} & \underline{.224} & .095 & .480 & .452 & .198 & .614 & .554 & \underline{.229} & 1 \\
 & 192 & .317 & .391 & .161 & \underline{.100} & \underline{.237} & \underline{.097} & .209 & \underline{.358} & \underline{.147} & \underline{.152} & \bf .231 & \bf .106 & \underline{.137} & \underline{.274} & \underline{.114} & .492 & .451 & .191 & .851 & .691 & .280 & 2 \\
 & 336 & .357 & .419 & \underline{.173} & \underline{.129} & \underline{.274} & \underline{.111} & .386 & .495 & .206 & \bf .148 & \underline{.233} & \bf .106 & \underline{.209} & \underline{.339} & \underline{.140} & .411 & .413 & .170 & .948 & .757 & .306 & 2 \\
 & 720 & .459 & .490 & .204 & \bf .180 & \bf .328 & \bf .132 & \underline{.749} & \underline{.736} & \underline{.329} & \underline{.170} & \underline{.255} & \underline{.116} & .344 & .449 & .184 & .498 & \underline{.454} & \underline{.189} & .988 & .799 & .323 & 3 \\
\cmidrule(lr){1-24}
\multirow{4}{*}{\rotatebox{90}{DualOrder}}
 & 96 & .298 & .380 & .159 & \bf .065 & \bf .188 & \bf .078 & .101 & \underline{.245} & \underline{.102} & \underline{.157} & \underline{.235} & \underline{.109} & \bf .093 & \underline{.224} & .095 & .481 & .441 & .194 & .615 & .551 & \underline{.229} & 4 \\
 & 192 & .318 & .392 & .162 & \bf .099 & \bf .236 & \bf .096 & .212 & \underline{.358} & .148 & .155 & \underline{.233} & \underline{.108} & .138 & .275 & \underline{.114} & \bf .456 & \bf .436 & \bf .184 & .846 & .681 & \underline{.278} & 6 \\
 & 336 & \underline{.354} & \underline{.418} & \underline{.173} & .130 & .275 & \underline{.111} & \bf .345 & \bf .469 & \bf .193 & \underline{.152} & .235 & \underline{.108} & .218 & .350 & .145 & .405 & .403 & .166 & .936 & \underline{.745} & \underline{.303} & 3 \\
 & 720 & .449 & .487 & .202 & .190 & .336 & .137 & .953 & .816 & .377 & \bf .168 & .257 & .117 & .358 & .462 & .190 & .514 & .464 & .194 & .994 & .791 & \underline{.317} & 1 \\
\cmidrule(lr){1-24}
\multirow{4}{*}{\rotatebox{90}{Concat}}
 & 96 & .287 & .374 & .156 & .072 & .198 & .084 & .108 & .255 & .108 & \underline{.157} & .238 & .110 & \bf .093 & \bf .222 & \underline{.094} & .470 & .437 & .193 & \underline{.606} & .556 & .236 & 2 \\
 & 192 & \underline{.312} & \underline{.388} & .161 & .103 & .240 & .099 & .219 & .370 & .154 & .156 & .241 & .110 & .138 & .275 & .115 & \underline{.457} & \underline{.440} & \underline{.188} & \underline{.819} & \underline{.678} & .283 & 0 \\
 & 336 & .362 & .421 & .174 & .137 & .283 & .115 & .413 & .507 & .212 & .160 & .251 & .116 & .213 & .343 & .142 & .411 & .407 & .169 & \underline{.924} & .755 & .306 & 0 \\
 & 720 & .466 & .496 & .207 & .189 & .335 & .136 & 1.02 & .844 & .384 & .185 & .274 & .126 & .345 & .449 & .184 & .491 & .457 & .190 & 1.00 & .801 & .324 & 0 \\
\bottomrule
\end{tabular}
}
\end{table}

\section{Differencing Order Ablation}
\label{sec:appendix-fp}

The default forward process uses $\mathbf{D}_2$, whose first row removes the absolute level, second row applies first-order differencing, and remaining rows apply second-order differencing. To validate this choice, we test two alternative constructions:
\begin{itemize}[leftmargin=1.5em,itemsep=2pt]
\item $\mathbf{D}_1$: all rows use first-order differencing $[-1,\,1]$ (captures velocity only).
\item $\mathbf{D}_3$: rows~1--2 same as $\mathbf{D}_2$; rows~3+ use third-order differencing $[-1,\,3,\,-3,\,1]$ (adds jerk information).
\end{itemize}
All three orders use the \textsc{DiffDiff} conditioning pathway with horizon-adaptive $\lambda$.
Table~\ref{tab:diff-order} reports results on the seven main benchmarks across four prediction horizons.

\begin{table}[t]
\centering
\caption{Differencing order ablation. $\mathbf{D}_k$ denotes the order of the differencing operator in the forward process, with $\mathbf{D}_2$ as the default used by \textsc{DiffDiff}. Within each column the best value across orders is in \textbf{bold}. The rightmost column reports per-row wins.}
\label{tab:diff-order}
\footnotesize
\setlength{\tabcolsep}{2.0pt}
\resizebox{\textwidth}{!}{
\begin{tabular}{c|c| ccc| ccc| ccc| ccc| ccc| ccc| ccc| |c}
\toprule
\multirow{2}{*}{Order} & \multirow{2}{*}{$H$}
 & \multicolumn{3}{c|}{Electricity}
 & \multicolumn{3}{c|}{ETTm2}
 & \multicolumn{3}{c|}{Exchange}
 & \multicolumn{3}{c|}{Traffic}
 & \multicolumn{3}{c|}{Weather}
 & \multicolumn{3}{c|}{Solar}
 & \multicolumn{3}{c||}{Wind}
 & \multirow{2}{*}{Wins} \\
\cmidrule(lr){3-5} \cmidrule(lr){6-8} \cmidrule(lr){9-11} \cmidrule(lr){12-14} \cmidrule(lr){15-17} \cmidrule(lr){18-20} \cmidrule(lr){21-23}
 & & MSE & MAE & CRPS & MSE & MAE & CRPS & MSE & MAE & CRPS & MSE & MAE & CRPS & MSE & MAE & CRPS & MSE & MAE & CRPS & MSE & MAE & CRPS & \\
\midrule
\multirow{4}{*}{\rotatebox{90}{$\mathbf{D}_1$}}
 & 96  & .283 & .370 & .154 & \bf .067 & \bf .189 & \bf .079 & .101 & \bf .246 & \bf .103 & .157 & .235 & \bf .108 & .094 & .224 & .095 & .465 & .437 & \bf .191 & .612 & .552 & .228 & 7 \\
 & 192 & .319 & .391 & .161 & .100 & .237 & \bf .097 & \bf .197 & \bf .348 & \bf .142 & .150 & .229 & .105 & .137 & .275 & .114 & .460 & .438 & .186 & .864 & .687 & .280 & 4 \\
 & 336 & .358 & .419 & .172 & \bf .128 & \bf .273 & .111 & \bf .388 & \bf .495 & \bf .207 & .147 & .231 & \bf .105 & .210 & .339 & .140 & .418 & .413 & .170 & .926 & .745 & .302 & 6 \\
 & 720 & .456 & .488 & .203 & \bf .182 & \bf .329 & \bf .133 & \bf .735 & \bf .728 & \bf .327 & .167 & .252 & \bf .114 & .339 & .443 & .182 & .492 & .454 & .188 & .978 & .794 & .321 & 7 \\
\cmidrule(lr){1-24}
\multirow{4}{*}{\rotatebox{90}{$\mathbf{D}_2$ (Ours)}}
 & 96  & \bf .269 & \bf .362 & \bf .150 & .068 & .190 & \bf .079 & .101 & \bf .246 & \bf .103 & \bf .156 & \bf .234 & \bf .108 & \bf .093 & \bf .222 & \bf .093 & \bf .463 & \bf .435 & \bf .191 & \bf .599 & \bf .547 & \bf .226 & 18 \\
 & 192 & .299 & .381 & .158 & .102 & .239 & .098 & .198 & .358 & .147 & .151 & .231 & .106 & .135 & .271 & \bf .112 & .483 & .448 & .191 & \bf .815 & \bf .675 & \bf .278 & 4 \\
 & 336 & \bf .349 & \bf .413 & \bf .171 & \bf .128 & \bf .273 & \bf .110 & \bf .388 & \bf .495 & \bf .207 & .148 & .232 & .106 & .202 & \bf .332 & \bf .137 & .389 & .395 & .163 & \bf .912 & \bf .742 & \bf .300 & 14 \\
 & 720 & \bf .435 & \bf .474 & \bf .196 & \bf .182 & .330 & \bf .133 & \bf .735 & \bf .728 & \bf .327 & .168 & .252 & \bf .114 & .328 & .434 & \bf .178 & .485 & .455 & .189 & \bf .966 & \bf .785 & \bf .317 & 13 \\
\cmidrule(lr){1-24}
\multirow{4}{*}{\rotatebox{90}{$\mathbf{D}_3$}}
 & 96  & .283 & .369 & .154 & \bf .067 & \bf .189 & \bf .079 & \bf .100 & \bf .246 & \bf .103 & \bf .156 & .235 & \bf .108 & .093 & .223 & .094 & .464 & .437 & \bf .191 & .610 & .551 & .228 & 10 \\
 & 192 & \bf .284 & \bf .376 & \bf .157 & \bf .099 & \bf .235 & .098 & .216 & .373 & .158 & \bf .137 & \bf .219 & \bf .101 & \bf .132 & \bf .268 & .113 & \bf .264 & \bf .314 & \bf .129 & .860 & .698 & .294 & 13 \\
 & 336 & .374 & .426 & .182 & .138 & .283 & .118 & .453 & .543 & .240 & \bf .132 & \bf .220 & .106 & \bf .199 & .337 & .142 & \bf .196 & \bf .284 & \bf .113 & 1.008 & .778 & .329 & 6 \\
 & 720 & 1.130 & .797 & .365 & .195 & .348 & .144 & 2.810 & 1.476 & .726 & \bf .151 & \bf .250 & .121 & \bf .313 & \bf .427 & .182 & \bf .242 & \bf .309 & \bf .122 & 1.183 & .868 & .382 & 7 \\
\bottomrule
\end{tabular}
}
\end{table}

\paragraph{Analysis.}
$\mathbf{D}_2$ achieves the most wins overall, supporting its choice as the default differencing operator.

$\mathbf{D}_1$ uses only first-order differences and loses to $\mathbf{D}_2$ on most settings, since $\mathbf{D}_2$'s additional curvature information provides a more discriminating inductive bias for the denoiser. It remains competitive only on benchmarks where the historical signal is dominated by smooth or strongly-trending dynamics.

$\mathbf{D}_3$ shows striking improvements on benchmarks dominated by periodic structure, where the third-order jerk information helps capture fast local change. However, its more aggressive differencing amplifies extrapolation errors on non-stationary series, producing severe long-horizon degradation on benchmarks where the future trajectory deviates strongly from a smooth continuation of the history. The strongly asymmetric behavior across data regimes makes $\mathbf{D}_3$ a less robust default than $\mathbf{D}_2$.

$\mathbf{D}_2$ therefore strikes a robust trade-off: enough differential structure to inform the denoiser without amplifying extrapolation errors on non-stationary regimes.

\section{Sensitivity Analysis}
\label{sec:appendix-sensitivity}

We conduct sensitivity analyses on four key hyperparameters of \textsc{DiffDiff}: the differencing strength $\lambda_{\max}$, the label window length $L_l$, the future loss weight $w_{\mathrm{fut}}$, and the number of DDIM sampling steps $S$.

\begin{table}[h]
\centering
\caption{Sensitivity to $\lambda_{\max}$.}
\label{tab:lambda-multih-v2}
\footnotesize
\setlength{\tabcolsep}{3.0pt}
\begin{tabular}{l|l|ccccc|ccccc}
\toprule
Dataset & H & \multicolumn{5}{c|}{MSE $\downarrow$} & \multicolumn{5}{c}{CRPS $\downarrow$} \\
 &  & 0.1 & 0.3 & 0.5 & 0.7 & 0.9 & 0.1 & 0.3 & 0.5 & 0.7 & 0.9 \\
\midrule
ETTm2 & 96 & .071 & .070 & .068 & .069 & .068 & .079 & .079 & .079 & .081 & .081 \\
 & 192 & .100 & .102 & .102 & .101 & .100 & .096 & .098 & .098 & .098 & .098 \\
 & 336 & .131 & .133 & .128 & .131 & .132 & .111 & .112 & .110 & .112 & .113 \\
 & 720 & .201 & .201 & .182 & .181 & .182 & .140 & .140 & .133 & .132 & .132 \\
\midrule
Exchange & 96 & .108 & .104 & .101 & .111 & .105 & .105 & .101 & .103 & .107 & .107 \\
 & 192 & .203 & .203 & .198 & .207 & .231 & .152 & .151 & .147 & .154 & .158 \\
 & 336 & .397 & .389 & .388 & .399 & .397 & .216 & .216 & .207 & .215 & .215 \\
 & 720 & .761 & .759 & .735 & .740 & .767 & .327 & .315 & .327 & .316 & .317 \\
\midrule
Electricity & 96 & .282 & .269 & .269 & .269 & .272 & .153 & .150 & .150 & .152 & .154 \\
 & 192 & .305 & .303 & .299 & .315 & .319 & .158 & .158 & .158 & .163 & .165 \\
 & 336 & .374 & .346 & .349 & .348 & .349 & .176 & .170 & .171 & .171 & .173 \\
 & 720 & .431 & .433 & .435 & .432 & .440 & .204 & .198 & .196 & .200 & .205 \\
\bottomrule
\end{tabular}
\end{table}

\begin{table}[h]
\centering
\caption{Sensitivity to label window length $L_l$.}
\label{tab:labellen-multih-v2}
\footnotesize
\setlength{\tabcolsep}{3.0pt}
\begin{tabular}{l|l|cccccc|cccccc}
\toprule
Dataset & H & \multicolumn{6}{c|}{MSE $\downarrow$} & \multicolumn{6}{c}{CRPS $\downarrow$} \\
 &  & 2 & 12 & 24 & 48 & 72 & 96 & 2 & 12 & 24 & 48 & 72 & 96 \\
\midrule
ETTm2 & 96 & .068 & .067 & .068 & .068 & .067 & .067 & .082 & .080 & .080 & .079 & .077 & .077 \\
 & 192 & .103 & .102 & .099 & .102 & .100 & .099 & .099 & .099 & .097 & .098 & .097 & .097 \\
 & 336 & .146 & .142 & .144 & .128 & .139 & .143 & .120 & .117 & .119 & .110 & .115 & .117 \\
 & 720 & .190 & .186 & .190 & .182 & .186 & .193 & .137 & .137 & .137 & .133 & .134 & .137 \\
\midrule
Exchange & 96 & .100 & .113 & .091 & .101 & .093 & .101 & .107 & .112 & .099 & .103 & .098 & .100 \\
 & 192 & .247 & .189 & .198 & .198 & .198 & .203 & .167 & .149 & .149 & .147 & .144 & .142 \\
 & 336 & .363 & .366 & .348 & .388 & .365 & .348 & .202 & .205 & .198 & .207 & .206 & .199 \\
 & 720 & .762 & .921 & 1.251 & .735 & .744 & .875 & .342 & .359 & .430 & .327 & .336 & .364 \\
\midrule
Electricity & 96 & .275 & .271 & .272 & .269 & .292 & .280 & .154 & .153 & .151 & .150 & .155 & .152 \\
 & 192 & .299 & .298 & .298 & .299 & .300 & .306 & .158 & .157 & .158 & .158 & .158 & .159 \\
 & 336 & .348 & .349 & .347 & .349 & .349 & .344 & .170 & .172 & .170 & .171 & .171 & .171 \\
 & 720 & .436 & .432 & .432 & .435 & .435 & .435 & .197 & .199 & .194 & .196 & .201 & .196 \\
\bottomrule
\end{tabular}
\end{table}

\begin{table}[h]
\centering
\caption{Sensitivity to future loss weight $w_{\mathrm{fut}}$.}
\label{tab:wt-multih-v2}
\footnotesize
\setlength{\tabcolsep}{3.0pt}
\begin{tabular}{l|l|cccc|cccc}
\toprule
Dataset & H & \multicolumn{4}{c|}{MSE $\downarrow$} & \multicolumn{4}{c}{CRPS $\downarrow$} \\
 &  & 1 & 3 & 5 & 10 & 1 & 3 & 5 & 10 \\
\midrule
ETTm2 & 96 & .068 & .068 & .068 & .068 & .079 & .079 & .079 & .079 \\
 & 192 & .102 & .102 & .102 & .103 & .098 & .098 & .098 & .098 \\
 & 336 & .128 & .131 & .128 & .133 & .110 & .112 & .110 & .112 \\
 & 720 & .183 & .184 & .182 & .184 & .133 & .134 & .133 & .134 \\
\midrule
Exchange & 96 & .112 & .108 & .101 & .103 & .106 & .104 & .103 & .103 \\
 & 192 & .265 & .226 & .198 & .198 & .169 & .159 & .147 & .150 \\
 & 336 & .368 & .368 & .388 & .414 & .203 & .206 & .207 & .213 \\
 & 720 & .749 & .778 & .735 & .762 & .347 & .326 & .327 & .326 \\
\midrule
Electricity & 96 & .285 & .274 & .269 & .277 & .156 & .152 & .150 & .152 \\
 & 192 & .314 & .310 & .299 & .299 & .163 & .160 & .158 & .158 \\
 & 336 & .348 & .351 & .349 & .344 & .172 & .172 & .171 & .171 \\
 & 720 & .455 & .429 & .435 & .432 & .204 & .197 & .196 & .196 \\
\bottomrule
\end{tabular}
\end{table}

\begin{table}[h]
\centering
\caption{Sensitivity to DDIM sampling steps $S$.}
\label{tab:steps-multih-v2}
\footnotesize
\setlength{\tabcolsep}{3.0pt}
\begin{tabular}{l|l|ccccc|ccccc}
\toprule
Dataset & H & \multicolumn{5}{c|}{MSE $\downarrow$} & \multicolumn{5}{c}{CRPS $\downarrow$} \\
 &  & 5 & 10 & 20 & 50 & 100 & 5 & 10 & 20 & 50 & 100 \\
\midrule
ETTm2 & 96 & .068 & .068 & .068 & .069 & .069 & .080 & .079 & .079 & .079 & .079 \\
 & 192 & .102 & .102 & .101 & .102 & .102 & .099 & .098 & .097 & .097 & .097 \\
 & 336 & .128 & .128 & .128 & .129 & .130 & .112 & .110 & .110 & .110 & .110 \\
 & 720 & .182 & .182 & .182 & .183 & .183 & .134 & .133 & .132 & .132 & .132 \\
\midrule
Exchange & 96 & .102 & .101 & .108 & .111 & .112 & .101 & .103 & .104 & .105 & .106 \\
 & 192 & .198 & .198 & .206 & .209 & .211 & .153 & .147 & .151 & .151 & .151 \\
 & 336 & .523 & .388 & .569 & .584 & .591 & .241 & .207 & .249 & .252 & .253 \\
 & 720 & .777 & .735 & .666 & .666 & .665 & .310 & .327 & .303 & .302 & .302 \\
\midrule
Electricity & 96 & .269 & .269 & .270 & .272 & .273 & .150 & .150 & .151 & .151 & .152 \\
 & 192 & .302 & .299 & .311 & .315 & .317 & .159 & .158 & .160 & .161 & .162 \\
 & 336 & .347 & .349 & .350 & .354 & .357 & .170 & .171 & .170 & .171 & .171 \\
 & 720 & .438 & .435 & .433 & .440 & .443 & .197 & .196 & .197 & .198 & .199 \\
\bottomrule
\end{tabular}
\end{table}

\section{Statistical Significance}
\label{sec:appendix-significance}

We report a paired Wilcoxon signed-rank test~\cite{demsar2006wilcoxon} on per-window CRPS comparing \textsc{DiffDiff} against strong baselines, namely, MA-TSD, NsDiff, and TMDM (Table~\ref{tab:significance}).

\begin{table}[htbp]
\centering
\caption{Paired Wilcoxon signed-rank test on per-window CRPS.}
\label{tab:significance}
\begin{tabular}{llccc}
\toprule
Dataset & $H$ & $p$ (vs.\ MA-TSD) & $p$ (vs.\ NsDiff) & $p$ (vs.\ TMDM) \\
\midrule
Electricity & 96  & $1.2\mathrm{e}{-18}$  & $3.0\mathrm{e}{-05}$ & $4.0\mathrm{e}{-03}$ \\
 & 192 & $5.7\mathrm{e}{-14}$  & $1.7\mathrm{e}{-08}$ & $4.8\mathrm{e}{-18}$ \\
 & 336 & $2.9\mathrm{e}{-50}$  & $2.3\mathrm{e}{-20}$ & $1.6\mathrm{e}{-18}$ \\
 & 720 & $1.5\mathrm{e}{-86}$  & $6.9\mathrm{e}{-01}$ & $7.3\mathrm{e}{-01}$ \\
\midrule
ETTm2 & 96 & $<10^{-300}$         & $1.3\mathrm{e}{-15}$ & $6.8\mathrm{e}{-02}$ \\
 & 192 & $3.2\mathrm{e}{-284}$    & $3.5\mathrm{e}{-15}$ & $1.7\mathrm{e}{-04}$ \\
 & 336 & $<10^{-300}$             & $1.9\mathrm{e}{-15}$ & $1.2\mathrm{e}{-02}$ \\
 & 720 & $<10^{-300}$             & $1.6\mathrm{e}{-11}$ & $3.3\mathrm{e}{-03}$ \\
\midrule
Exchange & 96 & $5.7\mathrm{e}{-39}$ & $1.2\mathrm{e}{-03}$ & $9.7\mathrm{e}{-01}$ \\
 & 192 & $2.2\mathrm{e}{-43}$       & $2.3\mathrm{e}{-02}$ & $3.9\mathrm{e}{-03}$ \\
 & 336 & $5.2\mathrm{e}{-02}$       & $5.1\mathrm{e}{-03}$ & $8.4\mathrm{e}{-03}$ \\
 & 720 & $1.7\mathrm{e}{-75}$       & $2.3\mathrm{e}{-02}$ & $1.0\mathrm{e}{+00}$ \\
\midrule
Traffic & 96 & $3.4\mathrm{e}{-98}$  & $1.5\mathrm{e}{-05}$ & $5.2\mathrm{e}{-17}$ \\
 & 192 & $1.8\mathrm{e}{-125}$       & $1.1\mathrm{e}{-03}$ & $8.8\mathrm{e}{-03}$ \\
 & 336 & $<10^{-300}$                & $6.2\mathrm{e}{-02}$ & $6.5\mathrm{e}{-02}$ \\
 & 720 & $<10^{-300}$                & $1.4\mathrm{e}{-10}$ & $7.6\mathrm{e}{-02}$ \\
\midrule
Weather & 96 & $1.5\mathrm{e}{-244}$ & $9.6\mathrm{e}{-11}$ & $3.8\mathrm{e}{-05}$ \\
 & 192 & $<10^{-300}$                & $1.3\mathrm{e}{-09}$ & $5.6\mathrm{e}{-02}$ \\
 & 336 & $<10^{-300}$                & $4.7\mathrm{e}{-02}$ & $4.4\mathrm{e}{-02}$ \\
 & 720 & $<10^{-300}$                & $9.9\mathrm{e}{-01}$ & $8.2\mathrm{e}{-01}$ \\
\midrule
Solar & 96 & $5.1\mathrm{e}{-97}$    & $6.8\mathrm{e}{-08}$ & $1.0\mathrm{e}{+00}$ \\
 & 192 & $1.8\mathrm{e}{-40}$        & $9.4\mathrm{e}{-01}$ & $1.0\mathrm{e}{+00}$ \\
 & 336 & $<10^{-300}$                & $1.0\mathrm{e}{+00}$ & $1.0\mathrm{e}{+00}$ \\
 & 720 & $<10^{-300}$                & $1.0\mathrm{e}{+00}$ & $3.6\mathrm{e}{-01}$ \\
\midrule
Wind & 96 & $2.2\mathrm{e}{-192}$    & $4.8\mathrm{e}{-02}$ & $7.2\mathrm{e}{-02}$ \\
 & 192 & $5.3\mathrm{e}{-226}$       & $1.5\mathrm{e}{-05}$ & $2.9\mathrm{e}{-01}$ \\
 & 336 & $4.3\mathrm{e}{-233}$       & $1.1\mathrm{e}{-01}$ & $1.0\mathrm{e}{+00}$ \\
 & 720 & $<10^{-300}$                & $1.8\mathrm{e}{-01}$ & $1.9\mathrm{e}{-02}$ \\
\bottomrule
\end{tabular}
\end{table}

{At a threshold of $p<0.05$, \textsc{DiffDiff} improves significantly over MA-TSD in 27 of 28 settings, over NsDiff in 20 of 28 settings, and over TMDM in 13 of 28 settings. The evidence is strongest against MA-TSD and NsDiff.}

\clearpage
\section{Computational Efficiency}
\label{sec:appendix-efficiency}

We report peak GPU memory and per-batch wall-clock for both training and inference. All measurements use ETTm2 with $H{=}720$ and batch size 32 on a single RTX 5090. Training memory and time cover one optimizer step of each method. Inference is unified across all methods to serial sampling: each method generates exactly $n_{\text{sample}}{=}100$ trajectories one trajectory at a time{, with the resulting comparison reported in Table~\ref{tab:efficiency}}.

\begin{table}[t]
\centering
\caption{Computation efficiency comparison. Bold face indicates the best result per row.}
\label{tab:efficiency}
\setlength{\tabcolsep}{4.0pt}
\begin{tabular}{lrrrrrr}
\toprule
& CSDI & TMDM & TimeDiff & NsDiff & MA-TSD & DiffDiff \\
\midrule
Mem.Train (MB)        & 181.33   & 4015.88  & 227.30   & 779.38   & \textbf{31.96}   & 38.57    \\
Mem.Inference (MB)    & \textbf{59.38} & 293.15   & 88.99    & 269.07   & 150.69   & 212.81   \\
Tim.Train (ms)        & 43.12    & 46.02    & 22.08    & 48.82    & \textbf{10.19}   & 12.95    \\
Tim.Inference (ms)    & 21284.91 & 10543.89 & 12671.42 & 3908.74  & 1705.27  & \textbf{1462.65} \\
CRPS                  & 1.12     & .150     & .182      & .159     & .145     & \textbf{.133} \\
\bottomrule
\end{tabular}
\end{table}

\section{{Distributional Fidelity Metrics}}
\label{sec:appendix-distribution-fidelity}

{We retain the evaluation notation $y_{i,t}$, $N$, $H$, and $K$ from Appendix~\ref{sec:appendix-metrics}, and denote the $k$-th generated value for window $i$ at position $t$ by $\tilde{y}_{i,t}^{(k)}$. Accordingly, $\mathbf{y}_i=[y_{i,1},\ldots,y_{i,H}]^\top$ and $\tilde{\mathbf{y}}_i^{(k)}=[\tilde{y}_{i,1}^{(k)},\ldots,\tilde{y}_{i,H}^{(k)}]^\top$ are the realized and generated trajectories. We use the metric-specific numerical floor $\varepsilon_{\mathrm{fid}}=10^{-12}$. Lower values indicate better fidelity for all three metrics.}

\paragraph{{Log-PSD RMSE.}}
{For a trajectory $\mathbf{v}\in\mathbb{R}^{H}$, let $\bar{v}=H^{-1}\sum_{t=1}^{H}v_t$ be its temporal mean, $\mathbf{1}\in\mathbb{R}^{H}$ the all-ones vector, and $\mathcal{F}$ the discrete Fourier transform. Its centered power at frequency bin $f$ is}
\begin{equation}
{
I_f(\mathbf{v})=
\left|\mathcal{F}(\mathbf{v}-\bar{v}\mathbf{1})_f\right|^2.
}
\label{eq:fid-periodogram}
\end{equation}
{Let $\Omega_H=\{0,\ldots,\lfloor H/2\rfloor\}$ denote the nonnegative Fourier bins. We denote the aggregate generated and realized spectra after unit-total-power normalization by $\widetilde{I}_f^{\mathrm{gen}}$ and $\widetilde{I}_f^{\mathrm{true}}$, respectively:}
\begin{equation}
{
\widetilde{I}_f^{\mathrm{gen}}=
\frac{\sum_{i=1}^{N}\sum_{k=1}^{K}I_f(\tilde{\mathbf{y}}_i^{(k)})}
{\sum_{j\in\Omega_H}\sum_{i=1}^{N}\sum_{k=1}^{K}I_j(\tilde{\mathbf{y}}_i^{(k)})}.
}
\label{eq:fid-generated-spectrum}
\end{equation}
\begin{equation}
{
\widetilde{I}_f^{\mathrm{true}}=
\frac{\sum_{i=1}^{N}I_f(\mathbf{y}_i)}
{\sum_{j\in\Omega_H}\sum_{i=1}^{N}I_j(\mathbf{y}_i)}.
}
\label{eq:fid-true-spectrum}
\end{equation}
{The root mean squared error (RMSE) between the two log spectra is}
\begin{equation}
{
\operatorname{RMSE}_{\mathrm{PSD}}=
\left[\frac{1}{|\Omega_H|}\sum_{f\in\Omega_H}
\left(
\log_{10}\max(\widetilde{I}_f^{\mathrm{gen}},\varepsilon_{\mathrm{fid}})-
\log_{10}\max(\widetilde{I}_f^{\mathrm{true}},\varepsilon_{\mathrm{fid}})
\right)^2\right]^{1/2}.
}
\label{eq:fid-log-psd}
\end{equation}

\paragraph{{Normalized $\mathbf{D}_2$-Wasserstein.}}
{For a trajectory $\mathbf{v}$ with value $v_t$ at forecast position $t$, we summarize its local changes by the mean squared second-difference energy}
\begin{equation}
{
\mathcal{E}_2(\mathbf{v})=
\frac{1}{H-2}\sum_{t=3}^{H}(v_t-2v_{t-1}+v_{t-2})^2.
}
\label{eq:fid-d2-energy}
\end{equation}
{The realized trajectories determine the normalization scale}
\begin{equation}
{
\overline{\mathcal{E}}_2^{\mathrm{true}}=
\max\left(
\frac{1}{N}\sum_{i=1}^{N}\mathcal{E}_2(\mathbf{y}_i),
\varepsilon_{\mathrm{fid}}
\right).
}
\label{eq:fid-d2-scale}
\end{equation}
{With $\delta_a$ denoting a point mass at $a$, $\widehat{\mathbb{P}}_2^{\mathrm{gen}}$ and $\widehat{\mathbb{P}}_2^{\mathrm{true}}$ denote the normalized empirical energy distributions of generated and realized trajectories:}
\begin{equation}
{
\widehat{\mathbb{P}}_2^{\mathrm{gen}}=
\frac{1}{NK}\sum_{i=1}^{N}\sum_{k=1}^{K}
\delta_{\mathcal{E}_2(\tilde{\mathbf{y}}_i^{(k)})/
\overline{\mathcal{E}}_2^{\mathrm{true}}}.
}
\label{eq:fid-generated-d2-distribution}
\end{equation}
\begin{equation}
{
\widehat{\mathbb{P}}_2^{\mathrm{true}}=
\frac{1}{N}\sum_{i=1}^{N}
\delta_{\mathcal{E}_2(\mathbf{y}_i)/
\overline{\mathcal{E}}_2^{\mathrm{true}}}.
}
\label{eq:fid-true-d2-distribution}
\end{equation}
{Using $W_1$ for the one-dimensional Wasserstein distance, their discrepancy is}
\begin{equation}
{
D_{\mathbf{D}_2}=
W_1\!\left(
\widehat{\mathbb{P}}_2^{\mathrm{gen}},
\widehat{\mathbb{P}}_2^{\mathrm{true}}
\right).
}
\label{eq:fid-d2-wasserstein}
\end{equation}
\paragraph{{Marginal QQ RMSE.}}
{Let $G_{\mathrm{gen}}$ and $G_{\mathrm{true}}$ be the empirical marginal distributions formed by pooling generated and realized values across windows and forecast positions, respectively, and let $G^{-1}(a)$ denote the quantile function of a distribution $G$ at level $a$. For the quantile grid $\mathcal{A}_{\mathrm{QQ}}=\{0.05,0.10,\ldots,0.95\}$, marginal quantile--quantile (QQ) error is normalized by $\operatorname{Std}(\{y_{i,t}\}_{i=1,t=1}^{N,H})$, the standard deviation of the pooled realized values:}
\begin{equation}
{
\operatorname{RMSE}_{\mathrm{QQ}}=
\frac{
\left[
\frac{1}{|\mathcal{A}_{\mathrm{QQ}}|}
\sum_{a\in\mathcal{A}_{\mathrm{QQ}}}
\left(G_{\mathrm{gen}}^{-1}(a)-G_{\mathrm{true}}^{-1}(a)\right)^2
\right]^{1/2}
}{
\max\left(
\operatorname{Std}\!\left(\{y_{i,t}\}_{i=1,t=1}^{N,H}\right),
\varepsilon_{\mathrm{fid}}
\right)
}.
}
\label{eq:fid-qq}
\end{equation}

\section{Limitations}
\label{sec:appendix-limitations}

{The fixed second-order differencing operator encodes a specific forecasting prior: smooth evolution is more strongly anchored by history than rapid local change. This prior may be less suitable when predictability is governed by irregular periodicities or cross-variable dependencies rather than a local frequency ordering. Future work can retain the locality and differencing bias of $\mathbf{D}_2$ while introducing learnable parameters that adapt the forward operator to dataset-specific predictability structure. The stage-adaptive gate also contributes less consistently than the forward process, with its clearest gains at the longest horizon, leaving room for conditioning mechanisms that adapt more reliably across horizons and data regimes.}

\end{document}